%% file: main.tex
\documentclass[journal]{IEEEtran} 

\usepackage[english]{babel}
\usepackage[utf8]{inputenc}
\usepackage[T1]{fontenc}
\usepackage{type1ec}
\usepackage{graphicx}
\usepackage{dsfont}
\usepackage{algorithm}
\usepackage{ bbold }
\usepackage{algorithmic}
\usepackage{amsmath}
\usepackage{multirow}
\usepackage[numbers,sort&compress]{natbib}
\usepackage{booktabs}
\usepackage{hyphenat}
\usepackage{url}

\usepackage{adjustbox}
\usepackage{caption}
\usepackage{subcaption}
\usepackage{amssymb}
\usepackage{multirow}

\usepackage{xcolor}

\usepackage{subfiles}

\newcommand{\R}{\mathds{R}}

\newcommand{\Ls}{\mathcal{L}}
\newcommand{\T}{\mathcal{T}}
\newcommand{\D}{\mathcal{D}}
\newcommand{\St}{\mathcal{S}}
\newcommand{\F}{\mathcal{F}}

\begin{document}

\title{Weakly Supervised Few-Shot Segmentation\\ Via Meta-Learning}

\author{Pedro H. T. Gama,~
        Hugo Oliveira,~
        José {Marcato~Junior},~
        Jefersson A. {dos~Santos}~
        
\thanks{Pedro~Gama and Jefersson~A.~dos~Santos are with the Department of Computer Science, Universidade Federal de Minas Gerais, Belo Horizonte, Brazil; phtg@dcc.ufmg.br}

\thanks{Hugo Oliveira is with the Institute of Mathematics and Statistics (IME), at University of S\~{a}o Paulo (USP), S\~{a}o Paulo, Brazil.}

\thanks{José Marcato Junior is with the Faculty of Engineering, Architecture and Urbanism and Geography, Federal University of Mato Grosso do Sul, Campo Grande, Brazil.} 

}

\maketitle

\begin{abstract}

Semantic segmentation is a classic computer vision task with multiple applications, which includes medical and remote sensing image analysis. 
Despite recent advances with deep-based approaches, labeling samples (pixels) for training models is laborious and, in some cases, unfeasible.
In this paper, we present two novel meta learning methods, named WeaSeL and ProtoSeg, for the few-shot semantic segmentation task with sparse annotations. 
We conducted extensive evaluation of the proposed methods in different applications (12 datasets) in medical imaging and agricultural remote sensing, which are very distinct fields of knowledge and usually subject to data scarcity. The results demonstrated the potential of our method, achieving suitable results for segmenting both coffee/orange crops and anatomical parts of the human body in comparison with full dense annotation.
\end{abstract}


\newcommand{\mamlmethod}{WeaSeL }
\newcommand{\protomethod}{ProtoSeg }
\newcommand{\cmamlmethod}{WeaSeL}
\newcommand{\cprotomethod}{ProtoSeg}

\input{introduction/intro.tex}
\input{relatedwork/related.tex}
\input{methodology/methodology.tex}
\input{experiments/experiments.tex}

\input{results/results.tex}
\input{conclusion/conclusion.tex}

\bibliographystyle{IEEEtranSN}
\bibliography{bibliography}

\end{document}


\title{Weakly Supervised Few-Shot Segmentation Via Meta-Learning: Supplementary Material}

\author{Pedro H. T. Gama \qquad Hugo Oliveira\\ José Marcato Junior \qquad Jefersson A. dos Santos
}
\date{2021}

\maketitle

\setcounter{table}{3}
\setcounter{figure}{16}
\setcounter{equation}{6}

\pagestyle{myheadings}

\section{miniUnet Architecture}

The network is comprised of three encoder blocks, a center block, three decoder blocks and a $1\times1$ convolution layer that works as a pixel-classification layer.
The network's blocks configuration can be seen in Table~\ref{tab:blocks}, where $C$ is the number of input channels obtained from the image domain (e.g. $C=3$ for RGB images or $C=1$ for radiology images) and the input/output features represent the number of feature dimensions that the input/output volume has (e.g. \textit{Encoder Block 1} receives an image of size $h\times w \times C$ and outputs a volume of size $\frac{h}{2}\times \frac{w}{2} \times 32$).
Similar to the classical UNet architecture \cite{ronneberger2015u}, skip connections are present in this model. This means that each decoder block receives as input the concatenation of the last block output and the corresponding encoder output. For instance, \textit{Decoder Block 1} receives as input the concatenation of the output volume of \textit{Decoder Block 2} and the output volume of \textit{Encoder Block 1}. 

\begin{figure}[h!]
    \centering
    \includegraphics[width=\columnwidth]{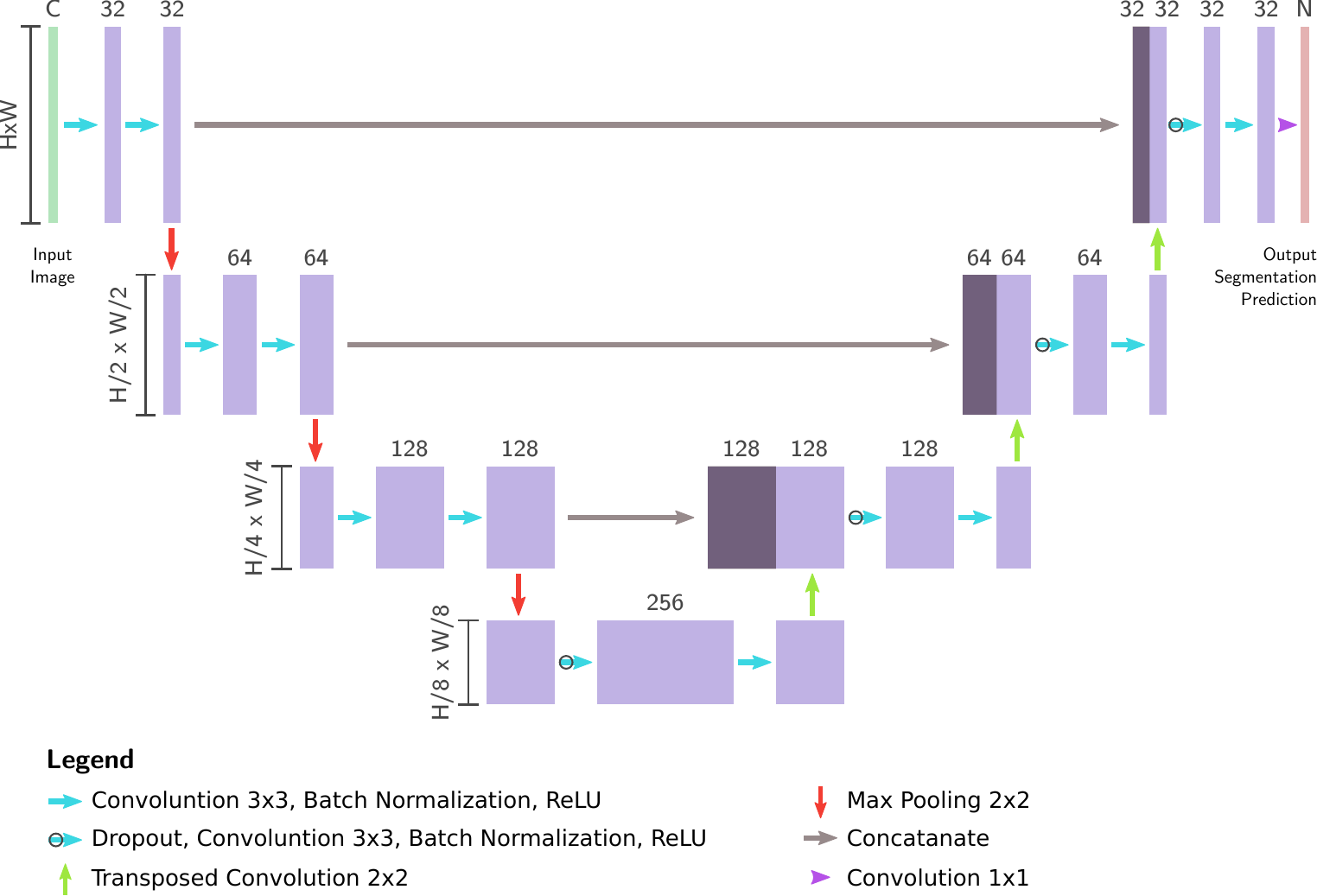}
    \caption{Illustration of the miniUNet architecture. The upper numbers represent the feature dimension of the volumes, while on the side is the spatial dimensions.}
    \label{fig:miniunet}
\end{figure}

Unlike the original architecture, we pad the images with zeros prior to the convolutions in order to preserve the spacial dimensions if the input. Hence, in the miniUnet architecture only pooling and transposed convolution operations affect the spatial dimensions of the volume during a forward pass in the network. A visualization of the miniUnet architecture can be seen in Figure~\ref{fig:miniunet}.

\begin{table}[h!]
    \centering
    \caption{Descriptions of the miniUNet blocks.}
    \label{tab:blocks}
    \setlength{\tabcolsep}{5pt}
    \begin{adjustbox}{width=0.55\columnwidth}
        \begin{tabular}{@{}ccc@{}}
            \toprule
            \textbf{Block Name} & \textit{Encoder Blocks (1, 2, 3)} & \textit{Center Block} \\ \midrule
            \textbf{Layers} &
              \begin{tabular}[c]{@{}c@{}}Conv $3\times3$\\ Bath Norm.\\ ReLU\\ Conv $3\times3$\\ Bath Norm.\\ ReLU\\ MaxPool $2\times2$\end{tabular} &
              \begin{tabular}[c]{@{}c@{}}Dropout\\ Conv $3\times3$\\ Bath Normalization\\ ReLU\\ Conv $3\times3$\\ Bath Normalization\\ ReLU\\ Transposed Conv $2\times2$\end{tabular} \\
              \midrule
            \multicolumn{1}{l}{\textbf{Input Feat./ Output Feat.}} & $C/32$, $32/64$, $64/128$         & $128/128$                  \\
            \toprule 
            \textbf{Block Name} & \textit{Decoder Blocks (3, 2)} & \textit{Decoder Block (1)} \\ \midrule
            \textbf{Layers} &
              \begin{tabular}[c]{@{}c@{}}Dropout\\ Conv $3\times3$\\ Bath Normalization\\ ReLU\\ Conv $3\times3$\\ Bath Normalization\\ ReLU\\ Transposed Conv $2\times2$\end{tabular} &
              \begin{tabular}[c]{@{}c@{}}Dropout\\ Conv $3\times3$\\ Bath Normalization\\ ReLU\\ Conv $3\times3$\\ Bath Normalization\\ ReLU\end{tabular} \\
              \midrule
            \multicolumn{1}{l}{\textbf{Input Feat./ Output Feat.}} & $256/64$, $128/32$                & $32/32$                    \\ \bottomrule
        \end{tabular}
    \end{adjustbox}
\end{table}

\section{Dataset Details}
\subsection{Medical Datasets}


    
    \textbf{1.A) JSRT Database}~\citep{JSRTshiraishi2000development} 
    is a collection 247 of chest radiographs initially proposed for lung nodules identification. Masks for lungs, clavicles and hearts structures were obtained from~\cite{JSRTvan_ginneken:2006-1223}. From each anatomical structure, a task is derived.
    All the images have a resolution of $2048\times2048$ and $12$ bit pixel resolution, and are gray scale.
    Examples are show in Figure 3(a).
    
    
    \textbf{1.B) Montgomery Dataset}~\citep{jaeger2014two}
    is a set of chests X-rays collected from patients in Montgomery County, Maryland, USA. There are 138 frontal X-rays, from which 58 are from cases of Tuberculosis - the initial use case of the dataset. Alongside with information of the patient, for each X-ray, there are binary masks for segmentation of each lung.
    All the X-rays are 12 bit gray scale images, and either have size $4020\times4892$ or $4892\times4020$. Examples are shown in Figure 3(b). 

    
    \textbf{1.C) Shenzhen Dataset}~\citep{jaeger2014two} 
    were publicized along with the Montogomery dataset. This set is comprised of chests X-rays collected from patients in Shenzhen, China. There are a total of 662 frontal X-rays, from which 336 are from cases of Tuberculosis. There is also binary masks for segmentation of each lung.
    The size of the X-rays vary, but average a $3000\times3000$ resolution, and are gray scale. Examples are shown in Figure 3(c).
    
    
    \textbf{1.D) NIH-labeled dataset}~\citep{NIHtang2019xlsor} 
    is a subset of the original NIH-labeled dataset~\citep{NIHwang2017chestx}. The original dataset is comprised of $108,948$ frontalview X-ray images of $32,717$ unique patients, and labeled with NLP for 14 different diseases.
    The subset used in this experiments will be named simply NIH-labeled or XLSor-NIH. This dataset proposed in \cite{NIHtang2019xlsor} is comprised of $100$ chest X-rays from the original NIH with manually annotated lung masks for these X-rays.
    All images have a spatial resolution of $512\times512$ and are gray scale. Examples can be seen in Figure 3(d).
    
    
    \textbf{1.E) OpenIST Chest X-Rays dataset}\footnote{\url{https://github.com/pi-null-mezon/OpenIST}} is a set of X-rays collected from the following original domain: \footnotesize{\url{http://www.chestx-ray.com/index.php/education/normal-cxr-module-train-your-eye\#!1}}. 
    \normalsize
    These images were used to train medical students in recognizing a normal X-ray. There are in total $225$ chest X-ray images, with binary masks for the lungs.
    The images are gray scale with 8-bit resolution and their sizes are not fixed. Examples in Figure 3(e).

    
    \textbf{1.F) LIDC-IDRI-DRR dataset}~\citep{LIDColiveira20203d} 
    is a dataset derived from LIDC~\citep{LIDCarmato2011lung}. This dataset is composed of flattened 2D Digitaly Reconstructed Radiographs (DRR) computed from chest CT-scans, as well as generated labels for the ribs.
    All the $835$ images in the dataset are gray scale, and have size $512\times512$. Examples of the scans and labels are show in Figure 3(f).
    
    
    \textbf{1.G) MIAS database}~\citep{MIASsuckling1994mammographic}
    is a collection of data from the Mammographic Image Analysis Society (MIAS), a research group in the UK with interest in mammograms. This dataset is composed of $322$ digitized mammograms, grayscaled and with a resolution of $1024\times1024$. The original dataset only provides labels of location and size of nodules in the images, but we had access to label masks that segment the pectoral muscles and the breast in each image.
    Example of samples are shown in Figure 3(g).
    
    
    \textbf{1.H) INbreast database}~\citep{moreira2012inbreast}
    is a collection of $410$ images collected from womans in the Breast Center located in the Centro Hospitalar de S. Joao [CHSJ], Porto, Portugal. From these $410$ images, only $~200$ are from Mediolateral Oblique (MLO) view (a side view of the breasts), and have labeled pectoral muscles for them. We also obtained labels for the breasts.
    All images are gray level with $14$ bit resolution and their sizes are either $3328\times4084$ or $2560\times3328$, depending on the patient. Examples can be seen in Figure 3(h).

    
    \textbf{1.I) Panoramic Dental X-rays}~\citep{PANORAMICabdi2015automatic}
    dataset is a set of panoramic dental X-rays of $116$ patients, taken at Noor Medical Imaging Center, Qom, Iran. The images were mannually segment by three specialists from which label masks for the mandibles were generated.
    All the images are gray scale and have a size of approximately $2900\times1250$ pixels. Examples of sample images and labels are presented in Figure 3(i).
    
    
    \textbf{1.J) IVisionLab Dental Images Dataset}~\citep{IVISIONsilva2018automatic}
    (or, simply, IVisionLab Dataset) is composed of a series of panoramic X-ray dental images. There is a total of $1500$ images with annotated teeth labels, with a variety of cases of dental problems and/or formations defects. 
    All the images are gray scale with dimensions of $2440\times1292$ pixels. Examples are shown in Figure 3(j).
    
\subsection{Remote Sensing Datasets}
    \textbf{2.A) Brazilian Coffee}~\citep{ferreira2018comparative,penatti2015deep}
     is a dataset comprised of 4 large satellite images from 4 municipalities of the state of Minas Gerais, Brazil - one satellite image for each county. These Counties being: Arceburgo, Guaranésia, Guaxupé and Montesanto. In the satellite images only three bands were considered, namely the Red, Green, and Near Infrared bands. Along these images, a binary ground truth label image is provided, with positive value representing a coffee crop and negative value representing the background. 
    For evaluation purpose, each of the images where cropped in non-overlapping patches of size $256\times256$, and only crops with a percentage of $25\%$ or more of pixels of coffee where maintained for training and validation of the models. Given that each county has distinct geographical features, which lead to coffee plantation distinctions, we consider each municipality a different task.
    Examples of patches of this dataset and their respective ground truths can be seen in Figure 4(a)-(d).
    
    \textbf{2.B) Orange Orchards}
    dataset is comprised of satellite images from an Orange Orchard located at the municipality of Ubirajara, São Paulo, Brazil. In the satellite images four bands are presented, Red, Green, Blue, and Near InfraRed, but only three of the bands were used: all but the Blue band. In addition to the images, annotation masks of the orange plantations were provided, making two classes, namely, Oranges and Background.
    Each of the satellite images where cropped in non-overlapping patches of size $384\times384$, and only crops with a percentage of $10\%$ or more of pixels of plantation where maintained for training and validation of the models. Since this dataset is focused in a single region and have only one interest class, we use it as a single task in our experiments.
    Examples of patches of this dataset and their respective ground truths can be seen in Figure 4(e).

\section{Additional Results}
In this section, we present extra results for the Section V-A, that were omitted for brevity.

\subsection{Extra Medical Tasks}
\begin{figure*}[h!]
    \centering
    \includegraphics[width=\textwidth]{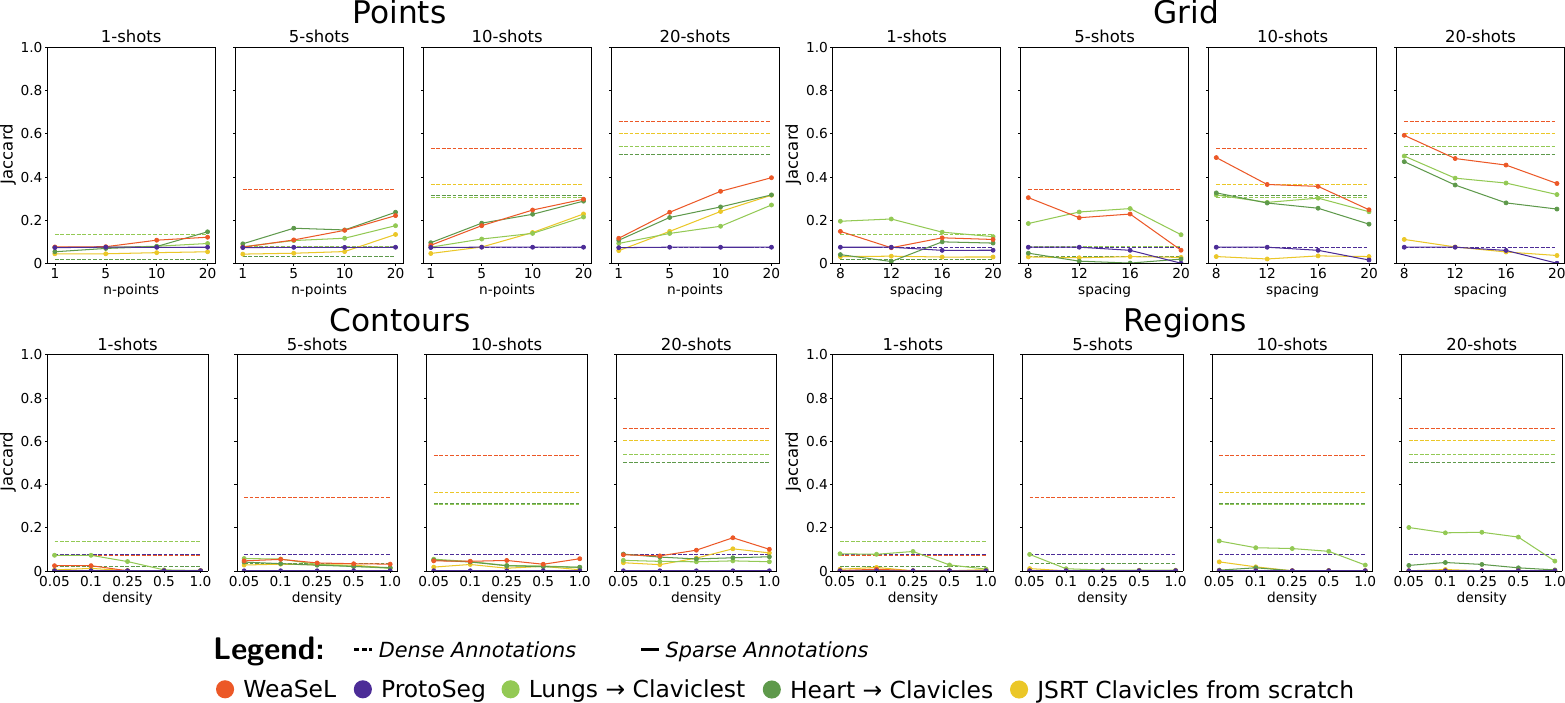}
    \caption{Jaccard score of experiments with \textit{JSRT Clavicles} task.}
    \label{fig:jsrt_clavicles_results}
\end{figure*}

\begin{figure*}[h!]
    \centering
    \includegraphics[width=\textwidth]{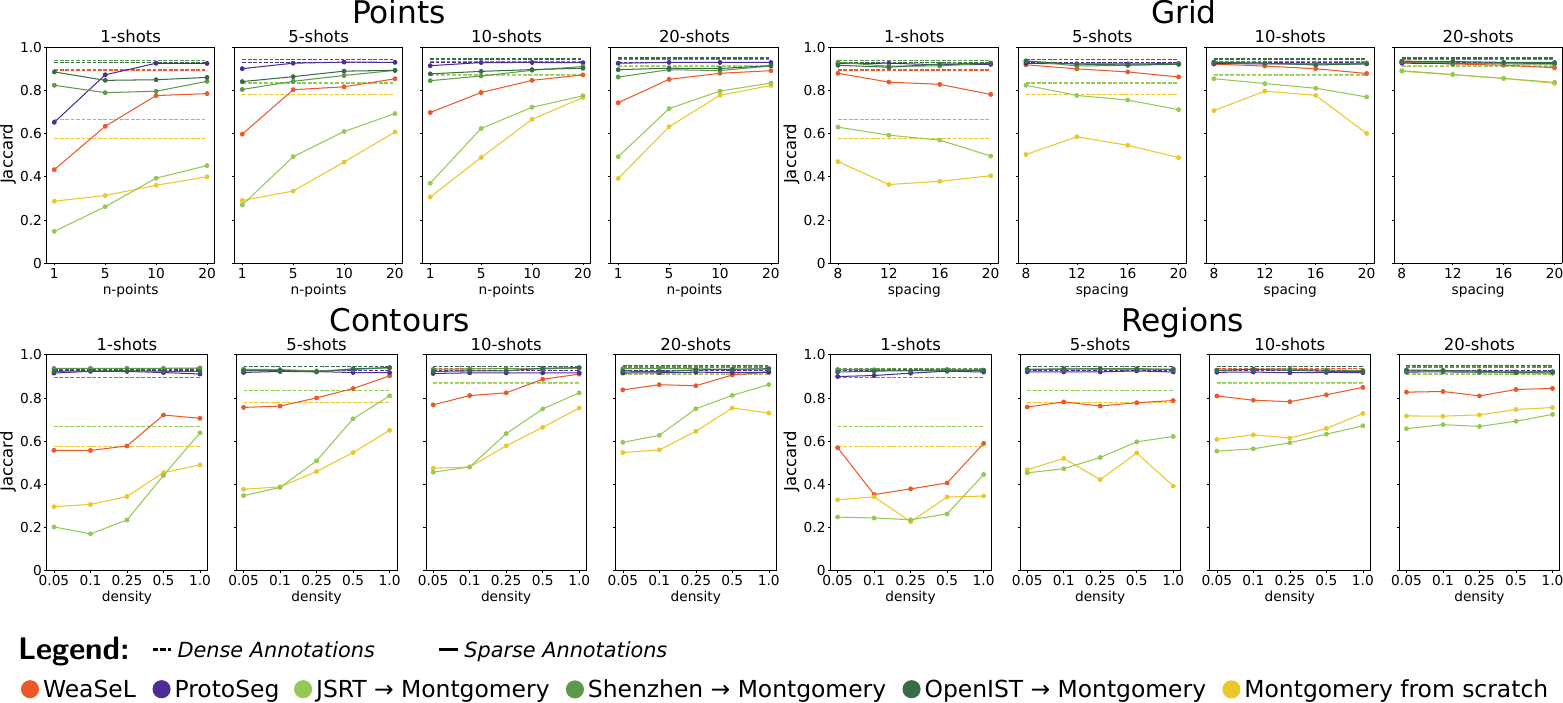}
    \caption{Jaccard score of experiments with \textit{Montgomery Lungs} task.}
    \label{fig:montgomery_lungs_results}
\end{figure*}

\begin{figure*}[h!]
    \centering
    \includegraphics[width=\textwidth]{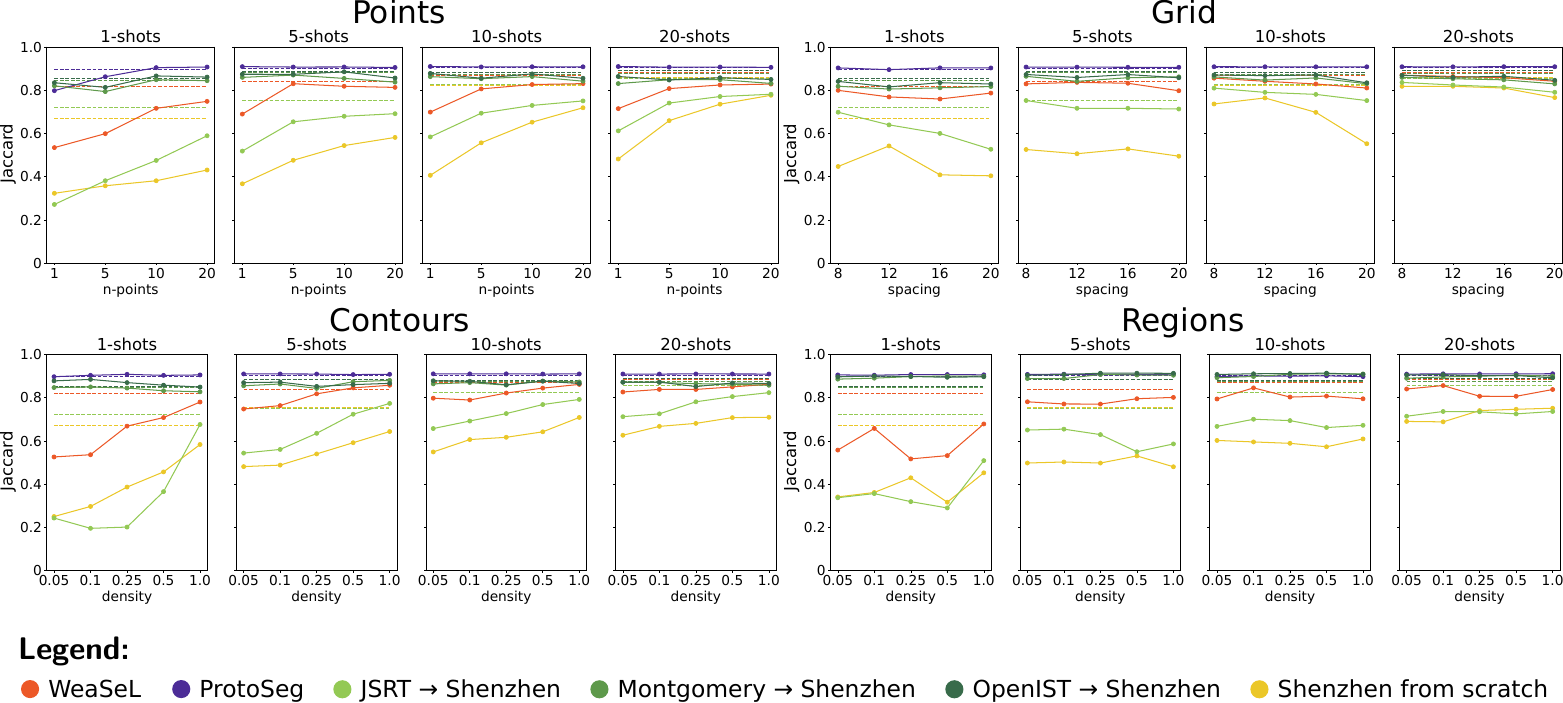}
    \caption{Jaccard score of experiments with \textit{Shenzhen Lungs} task.}
    \label{fig:shenzhen_lungs_results}
\end{figure*}

\begin{figure*}[h!]
    \centering
    \includegraphics[width=\textwidth]{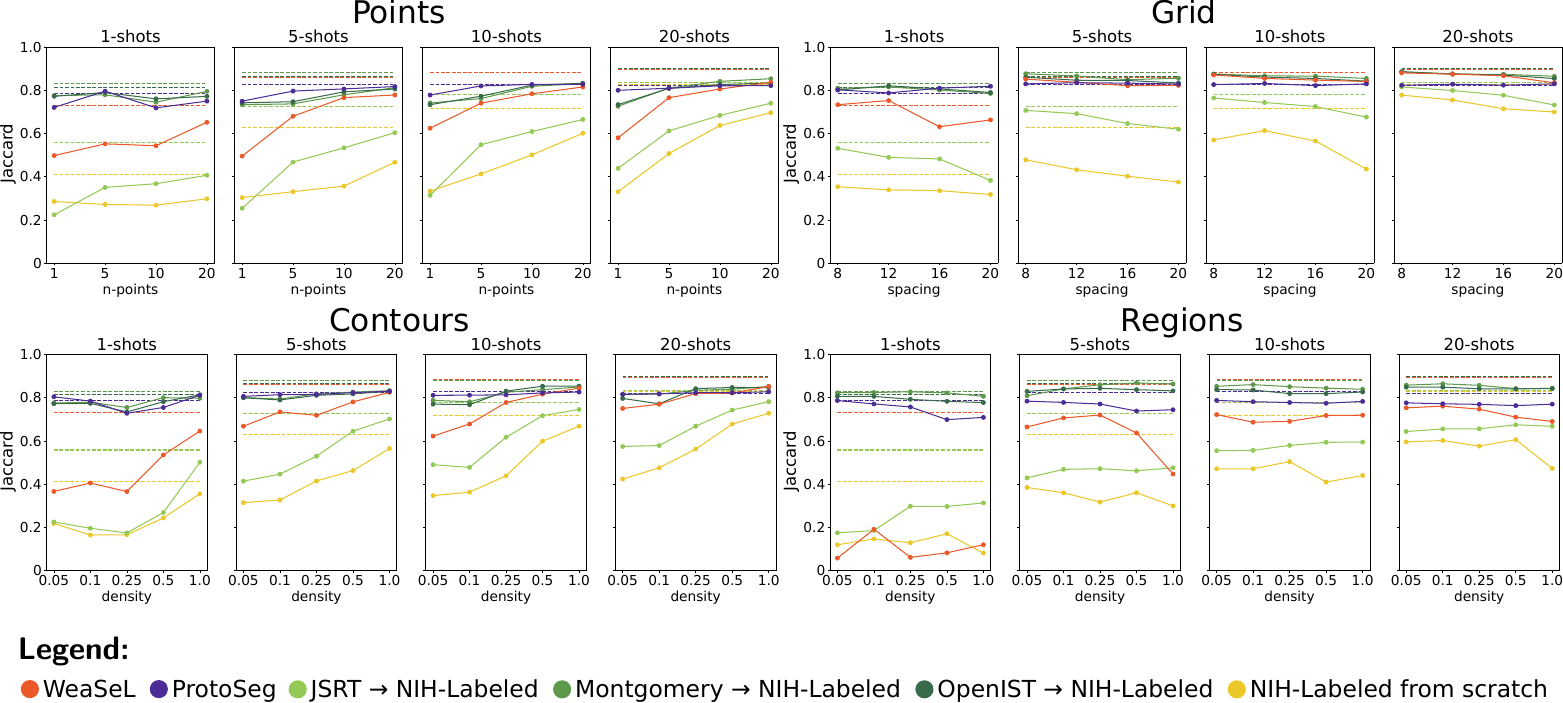}
    \caption{Jaccard score of experiments with \textit{NIH-labeled Lungs} task.}
    \label{fig:nih_labeled_lungs_results}
\end{figure*}

\begin{figure*}[h!]
    \centering
    \includegraphics[width=\textwidth]{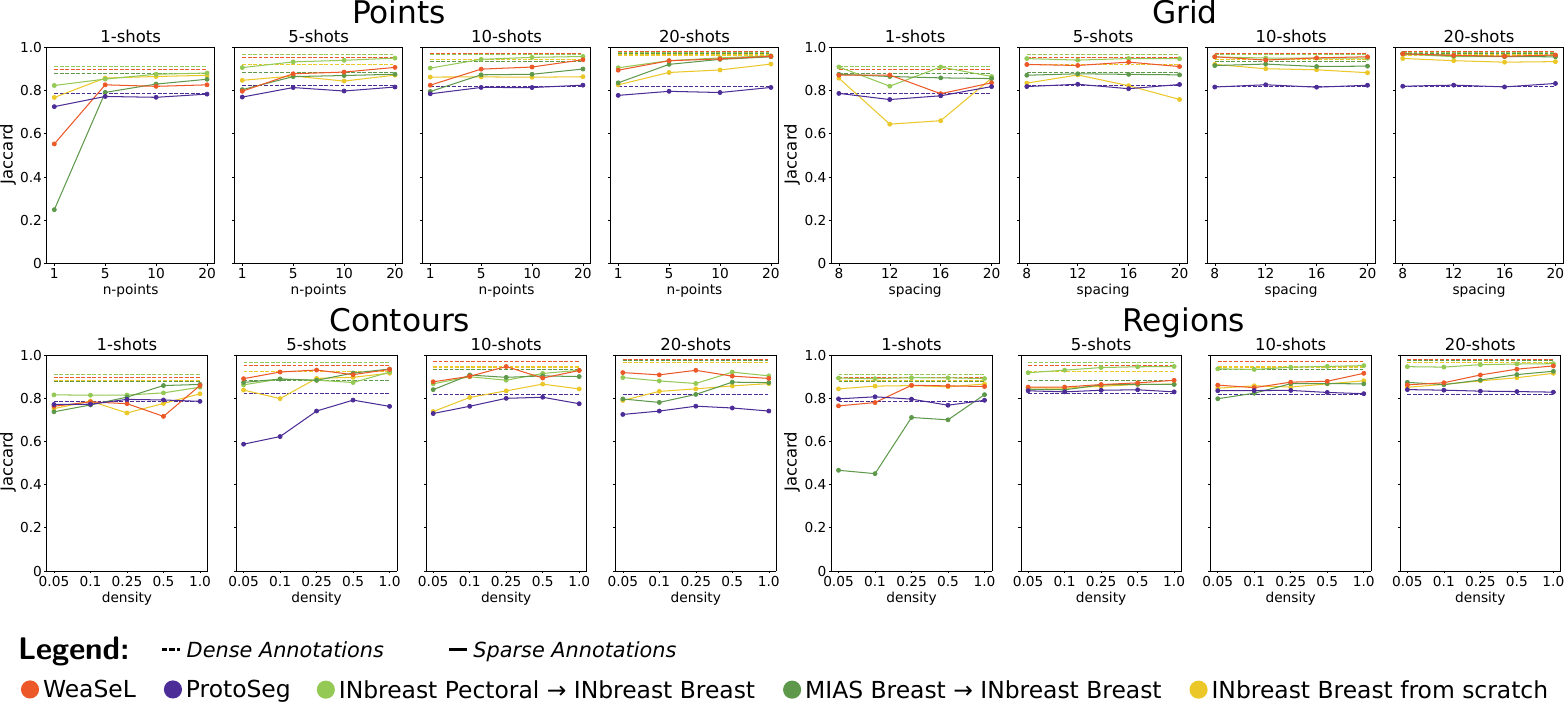}
    \caption{Jaccard score of experiments with \textit{INbreast Breast} task.}
    \label{fig:inbreast_breast_results}
\end{figure*}

\begin{figure*}[h!]
    \centering
    \includegraphics[width=\textwidth]{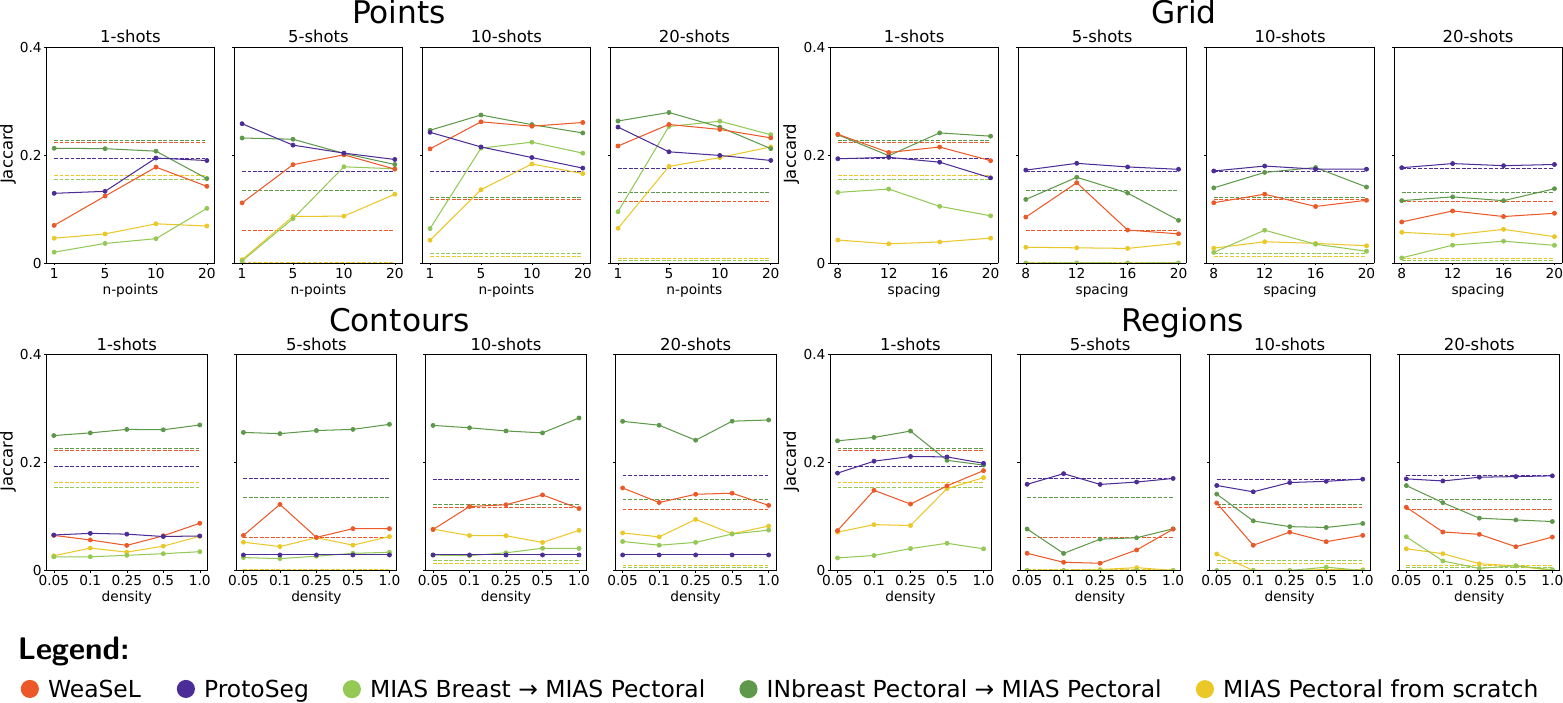}
    \caption{Jaccard score of experiments with \textit{MIAS Pectoral Muscle} task.}
    \label{fig:mias_pectoral_results}
\end{figure*}

\begin{figure*}[h!]
    \centering
    \includegraphics[width=\textwidth]{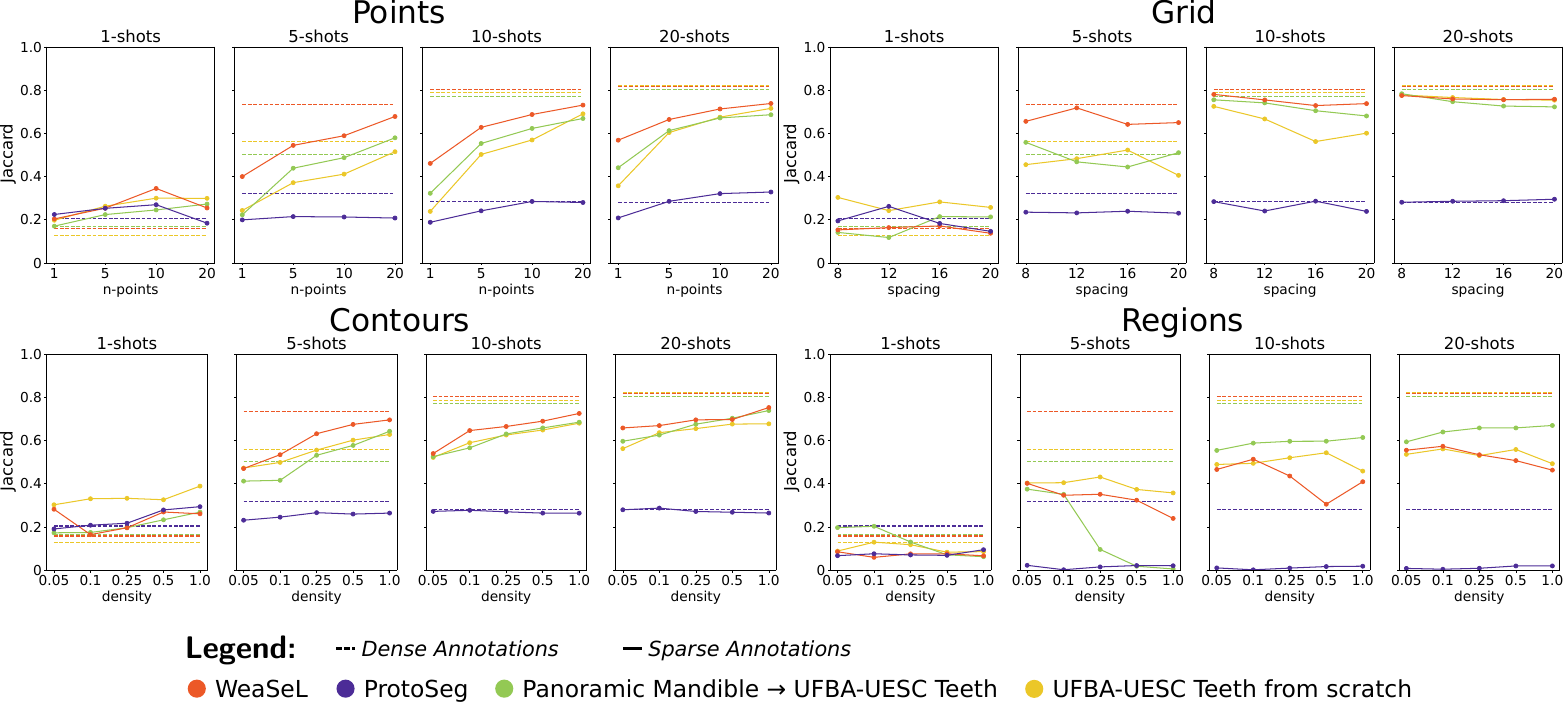}
    \caption{Jaccard score of experiments with \textit{IVisionLab Teeth} task.}
    \label{fig:ufba_teeth_results}
\end{figure*}

This section include the results of four omitted tasks of the Medical Experiments. These tasks are: \textit{JSRT Clavicles} (Figure~\ref{fig:jsrt_clavicles_results}), \textit{Montgomery Lungs} (Figure~\ref{fig:montgomery_lungs_results}),\textit{Shenzhen Lungs} (Figure~\ref{fig:shenzhen_lungs_results}), \textit{NIH-labeled Lungs} (Figure~\ref{fig:nih_labeled_lungs_results})\textit{MIAS Pectoral Muscle} (Figure~\ref{fig:mias_pectoral_results}), \textit{INbreast Breast} (Figure~\ref{fig:inbreast_breast_results}), and \textit{IVisionLab Teeth} (Figure~\ref{fig:ufba_teeth_results}).

\subsection{Extra Remote Sensing Tasks}

\begin{figure*}[h!]
    \centering
    \includegraphics[width=\textwidth]{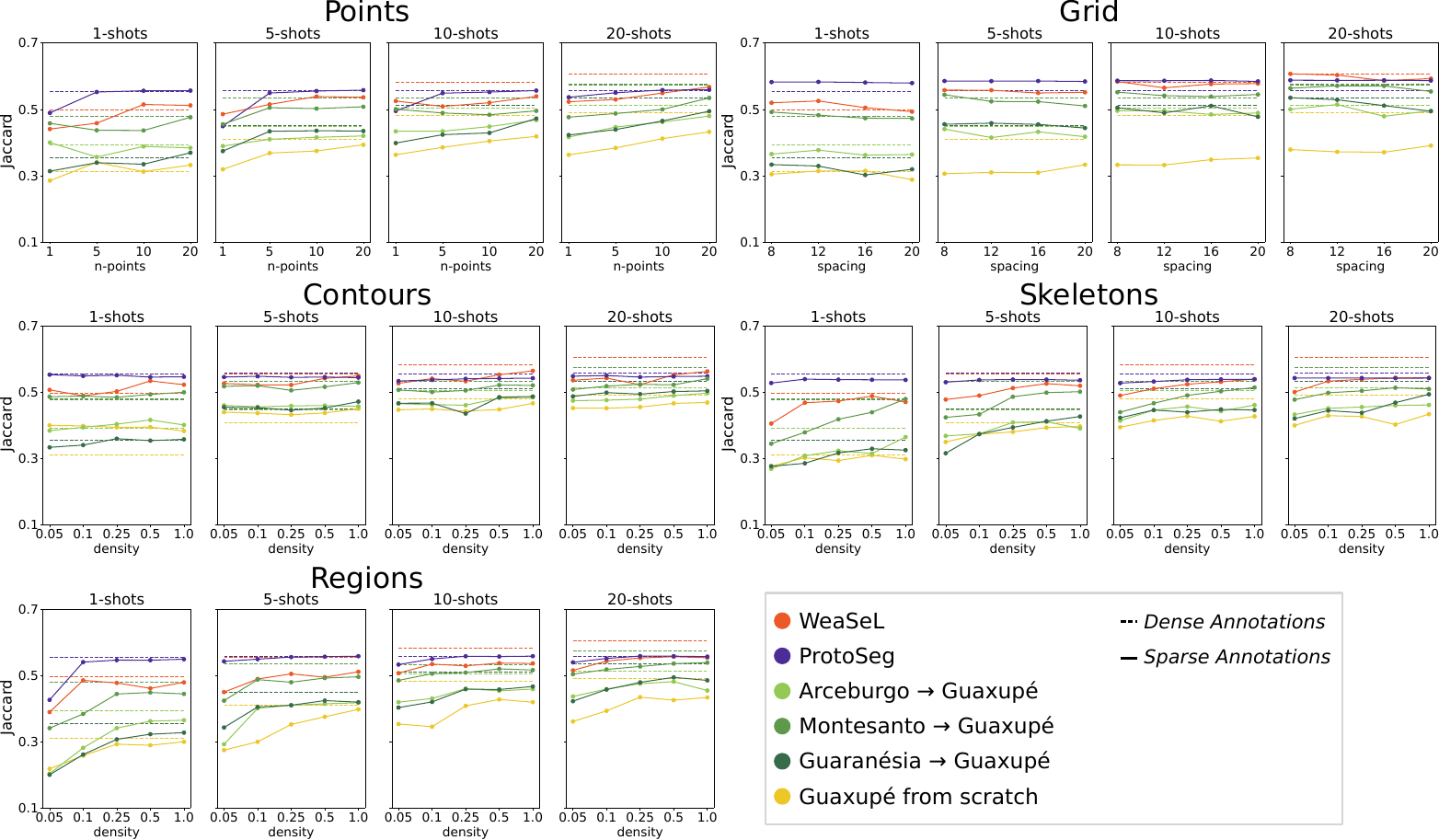}
    \caption{Jaccard score of experiments with \textit{Guaxupe Coffee} task.}
    \label{fig:guaxupe_results}
\end{figure*}

\begin{figure*}[h!]
    \centering
    \includegraphics[width=\textwidth]{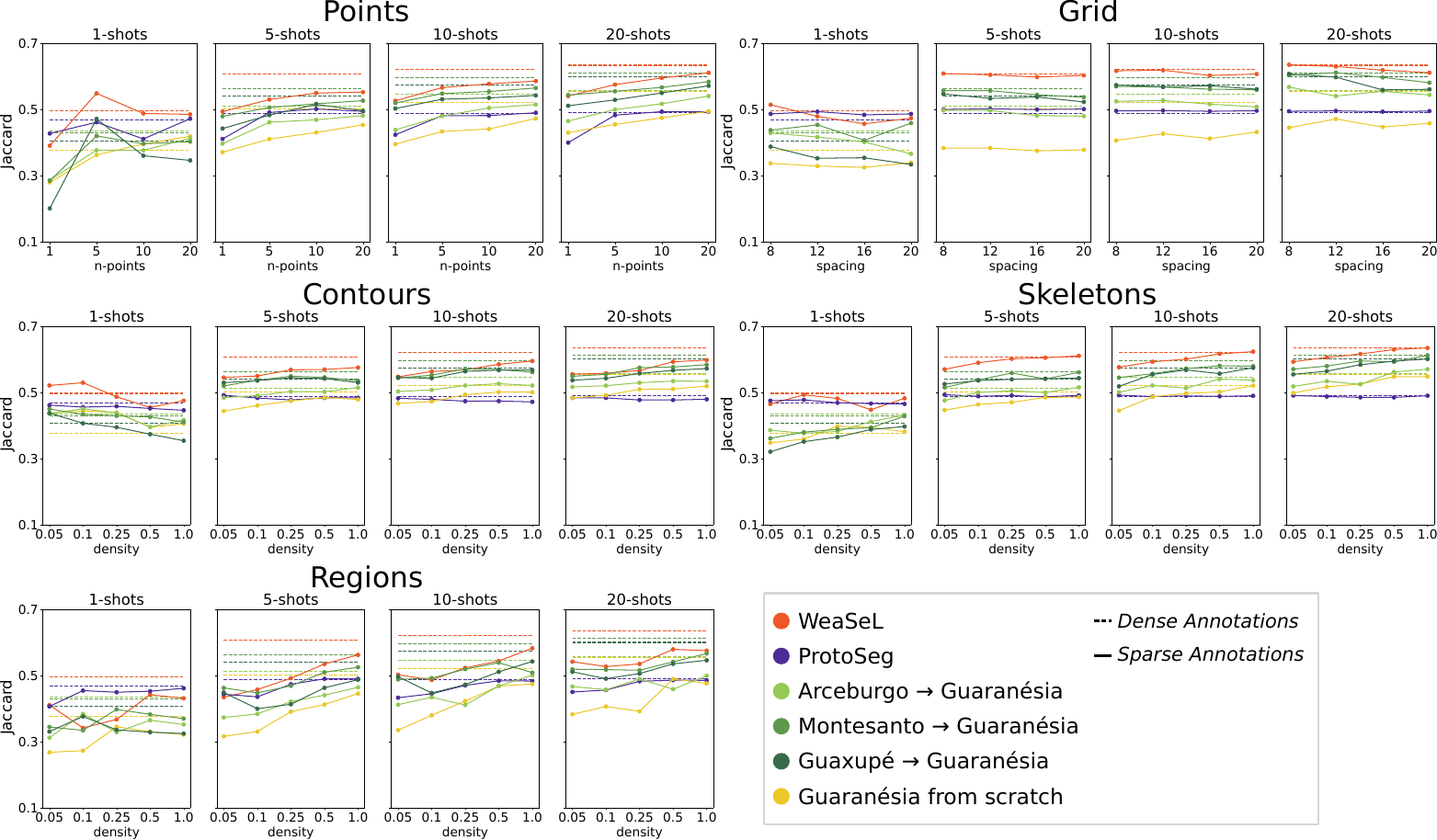}
    \caption{Jaccard score of experiments with \textit{Guaranesia Coffee} task.}
    \label{fig:guaranesia_results}
\end{figure*}

This section include the results of two omitted tasks of the Remote Sensing Experiments. The two tasks from the Brazilian Coffee dataset: \textit{Guaxupé Coffee} (Figure~\ref{fig:guaxupe_results}), and \textit{Guaranésia Coffee} (Figure~\ref{fig:guaranesia_results}).

\clearpage

\bibliographystyle{IEEEtranSN}
\small{
\bibliography{bibliography}
}

%% file: introduction/intro.tex
\section{Introduction}
\label{sec:intro}



Image segmentation is a classical computer vision problem where, given an image, a model is required to assign a class to every pixel, defining fine boundaries to the objects of interest that compose the image. It has applications in many scenarios, including medical image analysis \cite{ronneberger2015u,wang2018interactive}, remote sensing \cite{nogueira2015coffee,kataoka2016semantic}, and others.
State-of-the-art approaches to segmentation mostly use Deep Neural Networks (DNNs) methods, especially variations of Convolutional Neural Networks (CNNs). These approaches became popular after the work \citep{krizhevsky2012imagenet} and the advances of Graphical Processing Units (GPUs) that allowed the training of large complex models. 
The main limitation of current state-of-the-art deep models is the reliance on a large annotated training set, hampering the use of such models in more specific real-world scenarios out of the mainstream visual learning tasks. It is common for DNNs to present \textit{underfitting} or \textit{overfitting} \cite{Goodfellow-et-al-2016} problems when trained with a limited amount of data samples.

Common semantic segmentation methods rely on labels for all pixels in an image. From now on, this annotation strategy will be referred to as \textit{full/dense annotation}, being characterized by the highly laborious process required for producing such ground truths. The expensive process for producing \textit{dense annotations} is further aggravated in certain scenarios such as medical imaging or remote sensing, where usually only specialists are able to produce labels correctly. Thus \textit{sparse annotations} becomes an interesting solution, as they consist in only presenting a label for a small set of pixels of the image. 
This type of annotation reduces the time required to produce the labels for an image but, it can be challenging to train a model with such limitations on the amount of information available. Multiple methods \cite{lin2016scribblesup,vernaza2017learning,wang2018interactive} have successfully used sparse labels for image segmentation. 



Another strategy to reduce the cost of labeling datasets is to reduce the total number of images in it, and consequently the number of labeled images. 
Such scenarios with small dataset sizes are commonly known as few-shot and have recently gained the interest of the computer vision community. The few-shot learning literature contains a vast amount of works focused on image classification with notable examples \cite{vinyals2016matching,snell2017prototypical,finn2017model,raghu2019rapid}, although some methods for semantic segmentation have been proposed in recent years \cite{dong2018few,rakelly2018conditional,hu2019attention,zhang2019sgone,wang2019panet}. 

One methodology that has been successfully applied to few-shot problems in recent years is the meta-learning framework \cite{vinyals2016matching,snell2017prototypical,finn2017model}.
Normally understood as \textit{learning to learn}, meta-learning is an umbrella term for a collection of methods that improve the generalization of a learning algorithm through multiple multi-task learning episodes. A recent survey \cite{hospedales2020metalearning} formalizes the meta-learning framework and proposes different forms to categorize methods that use this approach. One can summarize meta-learning methods as an algorithm that learns a set of parameters $\omega$ called \textit{meta-knowledge}, trained using a distribution of tasks, such that $\omega$ generalizes well for the tasks in said distribution. The way a model achieves the training of the \textit{meta-knowledge} is used to group meta-learning methods into categories.

In this work, we extensively evaluated our previously proposed method \cmamlmethod~\cite{gama2021weakly} in a vast array of scenarios. Additionally, we introduced a fully novel semantic segmentation method (\cprotomethod), to problems with few-shot sparse annotated images. 
These two approaches are based on the meta-learning algorithms of Model-Agnostic Meta-Learning (MAML)~\cite{finn2017model}, and Prototypical Networks (ProtoNets)~\cite{snell2017prototypical}, respectively.

The main contributions of this work are:
(1) A novel meta-learning method for the problem of semantic segmentation with few-shot sparse annotated images;
(2) An extensive evaluation of our previous and new proposals in a large collection of tasks from Medical and Remote Sensing scenarios;
(3) A comparative analysis of five styles of sparse annotations named: Points, Grid, Contours, Skeletons, and Regions; and
(4) Two novel publicly available crop segmentation datasets with semantic labels for coffee and orange orchard crop regions. The coffee crop dataset has been previously published in previous works \cite{ferreira2018comparative,penatti2015deep}, but only for the task of patch classification. This work will be the first time this dataset is made fully publicly available with its semantic segmentation labels.

%% file: relatedwork/related.tex
\section{Related Work}
\label{chap:related}

\subsection{Weakly Supervised/Sparse Label Semantic Segmentation}\label{sec:weak_supervised}



Approaches to the problem of semantic segmentation with sparse labels can be mostly divided into two main groups: 1) methods that use the sparse labels without any kind of augmentation \cite{cciccek20163d,bokhorst2018learning,silvestri2018stereology,zhu2019pick}; and 2) strategies that try to reconstruct dense annotations from the sparse labels \cite{lin2016scribblesup,zhang2019sparse,bai2018recurrent,cai2018accurate}.

In the first group, \citet{cciccek20163d} and \citet{bokhorst2018learning} use a weighted loss, \citet{silvestri2018stereology} imply the use of padding in the sparse labels, and \citet{zhu2019pick} use a quality model to ensure a good segmentation based of the sparse annotation. 

\citet{lin2016scribblesup} are one of the first to use sparse labels for semantic segmentation. They use a label propagation scheme in conjunction with a FCN for segmentation. This propagation uses the scribble annotation provided and the prediction of the FCN network. They train their model by alternating which part is trained at each iteration. 

\citet{tajbakhsh2020embracing} present a thorough review of deep learning solutions to medical image segmentation problems. They include a section for segmentation with noisy/sparse labels, in which the methods belong to one of the two groups described previously. 
All the methods reviewed by \citet{tajbakhsh2020embracing} use a
selective loss. That is, a type of loss function that has different weights to unlabeled pixels/voxels and thus can ignore such pixels when the total cost is computed.

\subsection{Few-Shot Semantic Segmentation}\label{sec:few-shot_seg_semantic}

As in the few-shot classification problem, information from the support set has an important role in the semantic segmentation case. 
Multiple works try to insert this information directly into the model's processing flow. 

Many works use a two-branch structure, where one branch is responsible for extracting information from the support samples, which is fused into the other branch that processes the query images.
\citet{dong2018few} use a two-branch model, where one network produces prototypes of each class, similarly to ProtoNets. The first branch uses the support set and query image to produce prototypes, which are used for image classification in this branch. The encoded query image in the second branch is then fused with the prototypes to produce the probabilities maps.
\citet{hu2019attention} introduce a highly interconnected two-branch attention-based model. The attention modules receive query and support feature maps, and are present in multiple layers of the model.
\citet{zhang2019sgone} present another two-branch model. One branch, called \textit{guidance}, is used to extract feature vectors from both query and support images. They compute class prototypes using masked average pooling in the features from support images. A similarity map between the query features and prototypes is calculated and fused with the query features to compute the final prediction.

Other approaches use a single network to face the few-shot semantic segmentation problem.
\citet{wang2019panet} propose a direct adaptation of the Prototypical Networks. They use a CNN to produce feature vectors of the images in the support set and compute the prototypes for each class using masked average pooling, as~\cite{zhang2019sgone}. During the training they include an \textit{alignment} loss, where the prototypes are computed from the query image, and the support set is the segmentation target.
\citet{rakelly2018conditional} proposed the Guided Networks (Guided Nets), the first algorithm for few-shot sparse segmentation. Although it fuses information from the support set in query features, this model uses a single feature extraction network.
This network is a pre-trained CNN backbone that extracts features of the support set and the query image. The support features are averaged through a masked pooling using the sparse annotations provided for the images and further globally averaged across all the support images available. This single averaged support feature multiplies the query features, reweighting them. Then, these features are further processed by a small convolutional segmentation head that gives the final predictions. 

 
In Table~\ref{tab:literature}, we present a summary of the related works and how our proposed methods fit in the literature. 

\begin{table}[t!]
\caption{Summary of related work.}
\label{tab:literature}
\begin{adjustbox}{width=\columnwidth}
\begin{tabular}{@{}lccc@{}}
\toprule
\multicolumn{1}{l}{\textbf{Work}} & \textbf{Semantic Segmentation} & \textbf{Few-Shot} & \textbf{Sparse Annotations} \\
\toprule
\citet{lin2016scribblesup} & \checkmark &  & \checkmark \\
\citet{vernaza2017learning} & \checkmark &  & \checkmark \\
\citet{wang2018interactive} & \checkmark &  & \checkmark \\
\citet{zhang2019sparse} & \checkmark &  & \checkmark \\
\citet{bai2018recurrent} & \checkmark &  & \checkmark \\
\citet{cai2018accurate} & \checkmark &  & \checkmark \\
\citet{cciccek20163d} & \checkmark &  & \checkmark \\
\citet{bokhorst2018learning} & \checkmark &  & \checkmark \\
\citet{silvestri2018stereology} & \checkmark &  & \checkmark \\
\citet{zhu2019pick} & \checkmark &  & \checkmark \\
\citet{snell2017prototypical} &  & \checkmark &  \\
\citet{finn2017model} &  & \checkmark &  \\
\citet{dong2018few} & \checkmark & \checkmark &  \\
\citet{hu2019attention} & \checkmark & \checkmark &  \\
\citet{zhang2019sgone} & \checkmark & \checkmark &  \\
\citet{wang2019panet} & \checkmark & \checkmark &  \\
\citet{rakelly2018conditional} & \checkmark & \checkmark & \checkmark \\ \midrule
\mamlmethod (Ours) & \checkmark & \checkmark & \checkmark \\
\protomethod (Ours) & \checkmark & \checkmark & \checkmark \\ \bottomrule
\end{tabular}
\end{adjustbox}
\end{table}

%% file: methodology/methodology.tex
\section{Methodology}\label{chap:methodology}




\subsection{Problem Definition}
\label{sec:problem_def}


For our problem setup, we employ most of the definitions from \citet{gama2021weakly}. 

A dataset $\D$ is a set of pairs $(\mathbf{x}, \mathbf{y})$, where $\mathbf{x} \in \R^{H\times W \times B}$ is an image with dimensions $H\times W$ and $B$ bands/channels, and $\mathbf{y} \in \R^{H\times W}$ is the semantic label of the pixels in the image. This dataset is partitioned in two sets: $\D^{sup}$ (support set) and $\D^{qry}$ (query set), such that $\D^{sup} \cap \D^{qry} = \emptyset$.
Given a dataset $\D$ and target class $\T$, we define a segmentation task $\St$ as a tuple $\St = \{\D^{sup}, \D^{qry}, T\}$ (or, $\St = \{\D, \T\}$, for simplicity). 

A few-shot semantic segmentation task $\F$ is a specific type of segmentation task. It is also a tuple $\F = \{\D^{sup}, \D^{qry}, \T\}$, but the samples of $\D^{sup}$ have their labels sparsely annotated, and the labels in $\D^{qry}$ are absent or unknown. Moreover, the number of samples $k = |\D^{sup}|$ is a small number (e.g., $20$ or less); thus, we also call a few-shot task a $k$-shot task.

Finally, the problem of few-shot semantic segmentation with sparse labels is defined as follows. Given a few-shot task $\F$ and available segmentation tasks $\{\St_1 , \St_2 , \dots, \St_n \}$, we want to segment the images from the $\D^{qry}_{\F}$ using information from tasks $\St_i$, and information from the $\D^{sup}_{\F}$. 
Also, there is no information of the tasks $\F$ target objects, other than the sparse annotations of $\D^{sup}_{\F}$ samples. That is, no pair of image/semantic label of $\F$ is present in any task $\St_i$ in either $sup$ or $qry$ partition.

\subsection{Gradient-based Sparse Segmentation with \mamlmethod}
\label{sec:maml_seg}

\begin{figure}[h]
    \centering
    \begin{subfigure}{\columnwidth}
        \centering
        \includegraphics[width=\columnwidth]{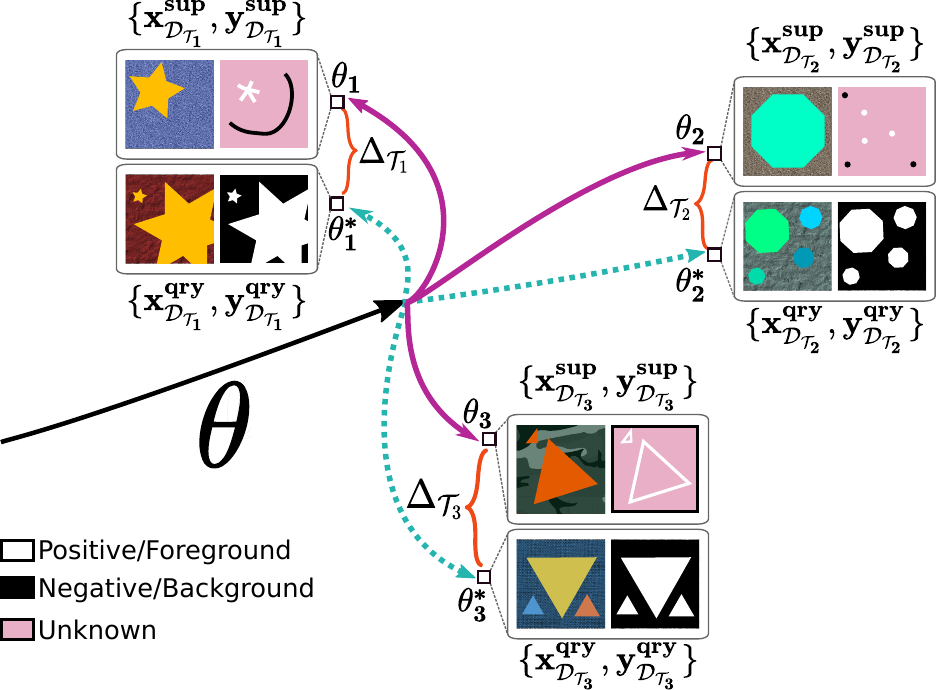}
        \caption{Visualization of the meta-training process. The global parameter $\theta$ is optimized for different tasks obtaining the parameters $\theta_i$ through optimization using the sparse labels from the tasks support sets. The $\theta_i^*$ is an optimal parameter that could be obtained if the model were trained with the query set samples and dense annotations, which are only used to compute the task outer loss. This hypothetical difference $\Delta$ between parameters is expected to be minimized during the meta-training, leading to a fast/better learner to the few-shot task.\\}
        \label{fig:maml_seg:meta-train}
    \end{subfigure}
    \begin{subfigure}{\columnwidth}
        \centering
        \includegraphics[width=\columnwidth]{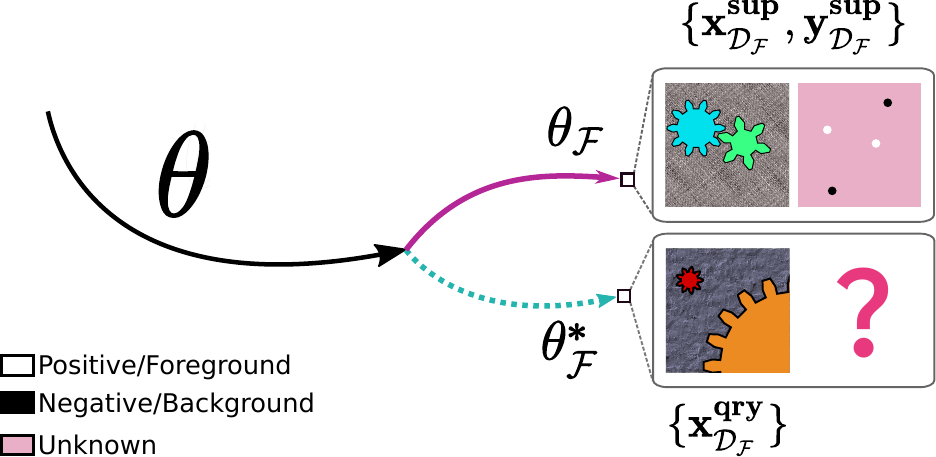}
        \caption{Illustration of the fine-tuning step. The meta-optimized $\theta$ is supervised trained with the sparse annotated samples of the few-shot support set. The labels of the query are unknown, i.e., not seen by the model.}
        \label{fig:maml_seg:finetune}
    \end{subfigure}
    \caption{Illustration of the WeaSeL method with toy examples in the meta-training/meta-test phase (a), and in the few-shot tuning phase (b).}
    \label{fig:maml_seg}
\end{figure}

We reintroduce our previously proposed method in \citet{gama2021weakly}. The \textbf{Wea}kly-supervised \textbf{Se}gmentation \textbf{L}earning (WeaSeL) is an adaptation of the supervised MAML algorithm \cite{finn2017model}, as depicted in Figure~\ref{fig:maml_seg}.

Our meta-tasks $\T_i \sim p(\T)$ are \textit{segmentation tasks} (i.e the set $\{\St_1 , \St_2 , \dots, \St_n \}$), as defined in Section~\ref{sec:problem_def}. We employ the Cross-Entropy loss ($\Ls_{CE}$) commonly used in segmentation tasks, defined for a single pixel as:
\begin{equation}\label{eq:cross_entropy}
    \Ls_{CE} = - \sum_{i}^{C} y_i \log f_{\theta}(x)_i \text{,}
\end{equation}
where C is the number of classes, $y_i$ is the true probability of a class $i$ for the pixel, and $f_{\theta}(x)_i$ is the predicted probability by the model $f_{\theta}$ for the class $i$ to the specific pixel. The loss in equation~\ref{eq:cross_entropy} is averaged over all pixels to produce the final loss for an image.


The meta-tasks have dense annotated samples. To train the model in a scenario similar to the target few-shot task, we simulate sparse annotations for the samples in the meta-tasks support set. That is, for all $\T_i \sim p(\T)$, the labels of samples in $\D^{sup}$ are randomly converted to a sparse version of themselves (this operation will be further discussed in Section~\ref{sec:type_of_annotations}). With this, we expect that the model learns to predict dense labels from sparse annotations and more easily adapts to few-shot tasks.

Given that the labels are sparse in the inner loop of meta-training, and during tuning in the few-shot task, we modify the classical Cross-Entropy to a Selective Cross-Entropy (SCE) loss as follows:
\begin{equation}\label{eq:selective_CE}
    \Ls_{SCE} = - \frac{1}{N} \sum_{j} \sum_{i}^{C} w_i y_i^j \log f_{\theta}(x)_i^j \text{,}
\end{equation}
where $j$ is a pixel, $N$ is the total number of labeled pixels, and $w_i$ is a indicator, where $w_i = 0$, if $i$ is an unknown label and $w_i=1$, otherwise. That is, $\Ls_{SCE}$ ignores pixels with unknown labels via the binary weight parameter, averaging the loss for all pixels with annotations.

Algorithm~\ref{alg:maml_gen} summarizes the meta-training procedure using the segmentation meta-task distribution $p(\T)$. In the inner loop, the loss is computed using the simulated sparse annotations of the support set of a task, and the outer loss using the dense labels of the query set of a task $\T_i$. 

After the meta-training phase, we adapt the model to the few-shot task, performing a simple fine-tuning with samples from the support set of the few-shot task $\F$. That is, we use pairs $(\mathbf{x}, \mathbf{y}) \in \D^{sup}_{\F}$ to train the model in a supervised manner using a Selective Cross-Entropy loss.

\begin{algorithm}
    \caption{Training algorithm for \mamlmethod.}
    \label{alg:maml_gen}
    \begin{algorithmic}
        \REQUIRE $p(\T)$: distributions over tasks
        \REQUIRE $\alpha, \beta$: step size hyperparameters
        
        \STATE Randomly initialize $\theta$
        \WHILE{not done}
            \STATE Sample batch of tasks $\T_i \sim p(\T)$
            \FORALL{$\T_i$}
                \STATE Sample batch of datapoints $S_i = \{(\mathbf{x}, \mathbf{y})\}$ from $\D^{sup}_{\T_i}$
                \STATE Compute $\nabla_\theta\Ls_{CE}(f_\theta)$ using $S_i$ and $\Ls_{SCE}$
                \STATE Update parameters: $\theta_i = \theta - \alpha\nabla_\theta\Ls_{SCE}(f_\theta)$
                \STATE Sample batch of datapoints $Q_i = \{(\mathbf{x}, \mathbf{y})\}$ from $\D^{qry}_{\T_i}$
            \ENDFOR
            \STATE Update $\theta \leftarrow \theta - \beta\nabla_\theta \sum_{\T_i}\Ls_{CE}(f_{\theta_i})$ using $Q_i$ and $\Ls_{CE}$
        \ENDWHILE
    \end{algorithmic}
\end{algorithm}


\subsection{Prototypical Seeds for Sparse Segmentation Via \protomethod}
\label{sec:prototypical_seg}


The proposed method for semantic segmentation based on the Prototypical Networks \cite{snell2017prototypical} is a straightforward adaptation of the original method. It uses the same premise of constructing a prototype vector to each class, with the distinction that prototypes are computed using the labeled pixels instead of whole image instances. 

Given a support set $S = \{(x_1, y_1), (x_2, y_2), \dots, (x_N, y_N)\}$, where $x_i \in \R^{H\times W \times B}$ is an image with height $H$, width $W$ and $B$ channels, and $y_i \in \R^{H\times W}$ is a label image with the semantic class of each pixel in $x_i$. Since $y_i$ can be sparse, the possible values of an pixel $j$ in ${y_i}$ are in the set $\{0, 1, 2, \dots, K\}$, where $K$ is the total of classes and $0$ represents the unknown class.

\begin{figure}
    \centering
    \includegraphics[width=\columnwidth]{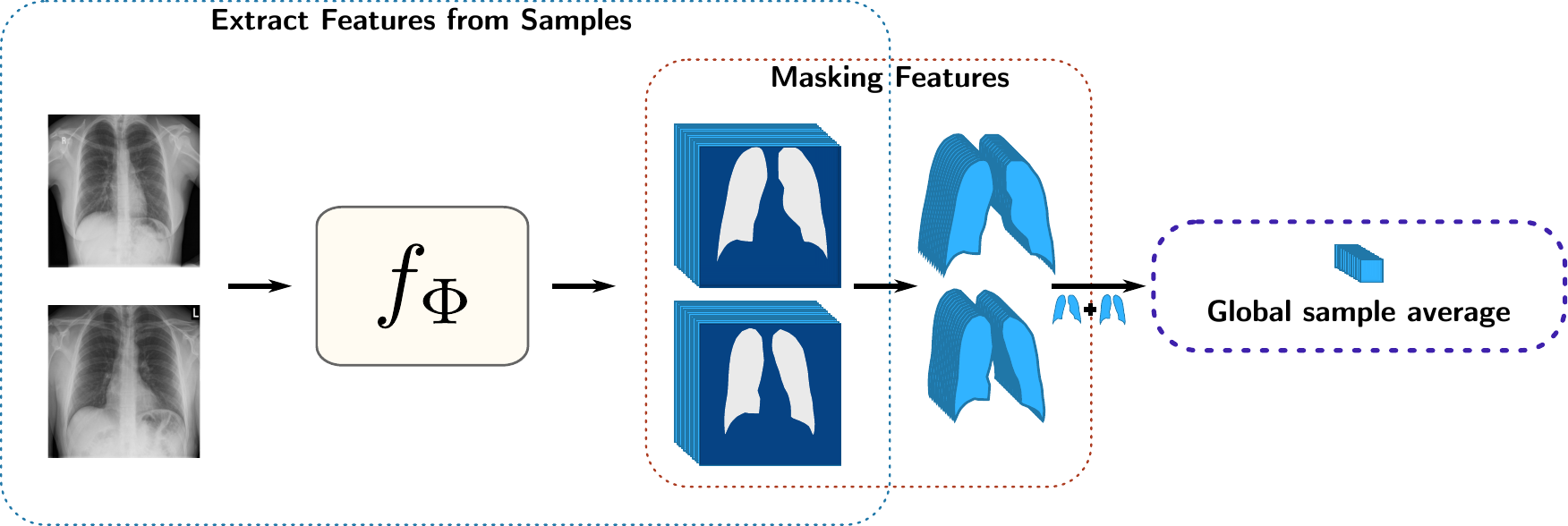}
    \caption{Illustration of our Average Pooling. After masking the features our process create the \textit{global sample average} by considering all pixels in the set.}
    \label{fig:mask_pooling}
\end{figure}

In our adaptation of ProtoNets, we define the $n$-dimensional prototype vector $\mathbf{c}_k$, of a class $k$ as:
\begin{equation}\label{eq:proposed_prototype}
    \mathbf{c}_k = \frac{1}{N_k} \sum_{(x_i, y_i) \in S} \sum_{j} [f_{\Phi} (x_i) \odot \mathbb{1}_k(y_i)]^{j} \text{,}
\end{equation}
where $f_{\Phi} : \R^{H\times W \times B} \rightarrow \R^{H\times W \times n}$ is our embedding function parametrized with $\Phi$ (a CNN), $\odot$ is point-wise multiplication, and $\mathbb{1}_k(y_i) \in \{0,1\}^{H\times W}$ is a mask matrix where each value is defined as
$$
\mathbb{1}_k^{j}(y_i) = 
\begin{cases}
1, \textnormal{ if } y_i^{j} = k \\
0, \textnormal{ otherwise}
\end{cases}    
$$
And $N_k = \sum_{y_i} \sum_j \mathbb{1}_k^{j}(y_i)$ is the total number of pixels of the class $k$, across all the support set $S$.
This means that our prototype vector $\mathbf{c}_k$ is the mean vector of all pixels of a class existent in the support set. 
This is similar to a masked average pooling, but considering all pixels globally, opposed to averaging for each sample and then averaging these pooled vectors. (See Figure~\ref{fig:mask_pooling}).

The inference is the same of the original Prototypical Networks, but applied to a pixel in the image. Formally, the probability of a pixel $j$ of a query image $\mathbf{q} \in \R^{H\times W \times B}$ belonging to a class $k$ is computed as follows:
\begin{equation}\label{eq:proposed_proto_dist}
    p_\Phi(y^{j} = k | \mathbf{q}) = \frac{\exp(-d(f_\Phi(\mathbf{q})^{j}, \mathbf{c}_k))}{\sum_{i} \exp(-d(f_\Phi(\mathbf{q})^{j}, \mathbf{c}_i))} \text{,}
\end{equation}
where $d$ is the squared euclidean distance: $d(\mathbf{u}, \mathbf{v}) = ||\mathbf{u} - \mathbf{v}||^2$.

Similar to the case of the \mamlmethod method, given the presence of unknown labeled pixels in training, we modify our loss function to ignore such pixels. We define our new loss function $J(\Phi)$ as follows:
\begin{equation}\label{eq:proto_loss}
    J(\Phi) = - \frac{1}{|Q|} \sum_{(\mathbf{x}, \mathbf{y}) \in Q} \sum_{j \in \mathbf{x}}\sum_{k=1}^{K} \log p_\Phi(\mathbf{y}^{j} = k | \mathbf{x}^{j}) \text{,}
\end{equation}
where $Q$ is the set of images used to compute the loss, $j$ is a pixel coordinate and $k$ represents a class. Note that $k$ starts from $1$, thus not considering the unknown class $k=0$. We use $p_\Phi$ as defined in equation~\ref{eq:proposed_proto_dist}. 

Given equations~\ref{eq:proposed_prototype} and \ref{eq:proposed_proto_dist}, the model $f_\Phi$ is trained using a episodic training strategy. 
This strategy resembles the training algorithm of WeaSeL and is presented in algorithm~\ref{alg:protonets}. It uses the same distribution over tasks $p(\T)$ as the first method, with the automatically generated sparse annotations of the meta-tasks in training. At each iteration, a batch of tasks is sampled, and for each task $\T_i$, a support set $S_i$ is constructed and used for training.

\begin{algorithm}
    \caption{Training algorithm for \protomethod.}
    \label{alg:protonets}
    \begin{algorithmic}
        \REQUIRE $p(\T)$: distributions over tasks
        
        \STATE randomly initialize $\Phi$
        \WHILE{not done}
            \STATE Sample batch of tasks $\T_i \sim p(\T)$
            \FORALL{$\T_i$}
                \STATE Sample a support set $S_i = \{(x_1, y_1), \dots, (x_N, y_N)\}$ from $\D^{sup}_{\T_i}$
                \STATE Compute $\mathbf{c}_k$ using $S_i$, for all $k$ using equation~\ref{eq:proposed_prototype}
                \STATE Sample query batch $Q_i = \{(\mathbf{x}_{1}, \mathbf{y}_{1}),  \dots, (\mathbf{x}_{b}, \mathbf{y}_{b})\}$ from $\D^{qry}_{\T_i}$
                \STATE Compute the loss $J(\Phi)$ as defined in equation~\ref{eq:proto_loss}, using $Q_i$.
                \STATE Update $\Phi$ using gradient descent and $\nabla J$
            \ENDFOR
        \ENDWHILE
    \end{algorithmic}
\end{algorithm}

%% file: experiments/experiments.tex
\section{Experimental Setup}\label{chap:exp_setup}

In this section, we present the configurations used for our experiments. In Section~\ref{sec:datasets}, we briefly present the datasets used. The evaluated sparse labels annotation styles are listed in Section~\ref{sec:type_of_annotations}. Next, in Section~\ref{sec:mini_unet}, we introduce the FCN architecture used, and in Section~\ref{sec:protocol} the baselines, protocol, and metrics are presented.

All the code for the experiments were written in the Python3 language. For the models, we use the Pytorch\footnote{\url{https://pytorch.org}} framework and the Torchmeta\footnote{\url{https://github.com/tristandeleu/pytorch-meta}} module. In relation to the machine, all the experiments were performed on Ubuntu SO, 64-bit Intel i9 7920X machine with 64GB of RAM memory, and a GeForce RTX 2080 TI/Titan XP GPU (only one GPU was used during the experiments).

\input{experiments/datasets}

\subsection{Types of sparse annotation}
\label{sec:type_of_annotations}


\begin{figure*}[h!]
    \centering
    \includegraphics{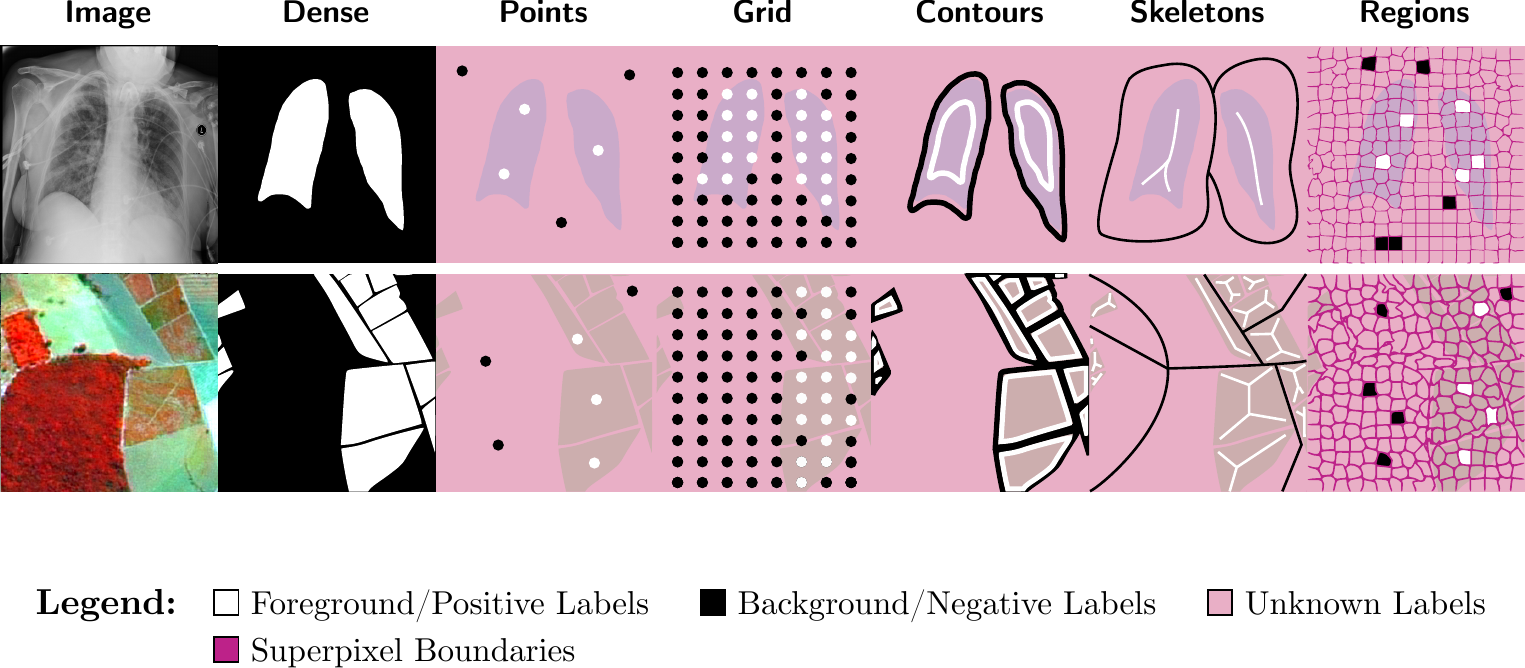}
    \caption{Illustration of the types of sparse annotations used. Annotations are illustrative and uspcaled to better visualization.}
    \label{fig:sparse_examples}
\end{figure*}

In the experiments we evaluate five types of sparse annotation, namely: \textit{points}, \textit{grid}, \textit{contours}, \textit{skeletons}, and \textit{regions}. As mentioned, we simulate these annotations from the original dense labels of an image. Visual examples of these annotations are show in Figure~\ref{fig:sparse_examples}.
We can describe each type of annotation and explain how the sparse annotations are generated as follow:

\renewcommand\theenumi{\textbf{\Roman{enumi}}}

\begin{enumerate}
     \item \textbf{Points:} It simulates an annotator alternately picking pixels from the foreground and background classes. We use a parameter $n$ and randomly choose $n$ pixels from the foreground and $n$ from the background. The remainder pixels are set as unknown.
   
    
    \item \textbf{Grid:} The annotator receives a pre-selected collection of pixels of the image, which are initially assumed to be from the background class. The pixels considered foreground should be annotated. These pre-selected pixels are disposed of in a grid pattern that was generated by using a parameter $s$. First, a random pixel $p_0$ is selected within the following rectangular region: \{upper left corner: $(0,0)$ and bottom right corner: $(s,s)$ \}. Afterward, a grid is created from this $p_0$ position with $s$ spacing horizontally and vertically. Pixels outside the grid are set as unknown.
    
    \item \textbf{Contours:} The annotator denotes the inner and outer boundaries of foreground objects. This style is useful for cases where a single connected foreground object is present. 
    We simulate these annotations by using morphological operations on the original binary dense labels. We used an erosion operation followed by a marching squares algorithm\footnote{\url{https://scikit-image.org/docs/0.8.0/api/skimage.measure.find_contours.html\#find-contours}} to find the inner contours. To the outer contours, we use a dilation operation on the original label mask and the same marching squares algorithm. Additionally, we use a parameter $d$ that determines the density of the sparse annotation. 
     
    \item \textbf{Skeleton-Based Scribble:} It resembles an annotator drawing a scribble roughly at the center of the foreground objects that more or less approximate the object form. The same process is applied to the background.
    These annotations are generated using the \textit{skeletonize algorithm}\footnote{\url{https://scikit-image.org/docs/0.8.0/api/skimage.morphology.html?highlight=skeletonize\#skeletonize}} in the binary dense label masks, which returns the skeletons of the foreground objects. The same process is applied to the negative dense label masks to obtain the skeleton of the background class. Dilation is applied to add thickness to the skeletons.
    We use a parameter $d$ to control the density of the annotation. We generate random binary blobs (using this\footnote{\url{https://scikit-image.org/docs/dev/api/skimage.data.html\#skimage.data.binary_blobs}} function) that occupy $d$ percentage of the image space and use them to mask the computed skeletons. 
    
    
    \item \textbf{Regions:} This type of annotation represents the process of an annotator appointing classes to \textit{pure superpixels}. We define a \textit{pure superpixel} as a usually small connected set of pixels with the same class. The annotator is provided with the superpixels of the image and then appoints the class of a subset of pure foreground and background superpixels.
    To generate these annotations, first, we compute the superpixels of the images using the SLIC algorithm~\citep{slic2012achanta} with empirically chosen parameters for each dataset. Once superpixels were computed,  we randomly selected a $d$ percentage of the superpixels for the foreground and a $d$ percentage of superpixels for the background.
\end{enumerate}

\subsection{miniUNet architecture}\label{sec:mini_unet}
The network model used in all experiments -- Baselines, \cmamlmethod, and \cprotomethod -- is a simplified version of the UNet architecture~\citep{ronneberger2015u}. 
We will call it miniUNet since it is a smaller version of the original network.
More information about the miniUNet architecture can be seen in the supplementary material of this manuscript.

In \cprotomethod, since we want to generate $n$-dimensional feature vectors, the last layer of the network is ignored and the output is gathered from the last decoder block. That is, the embedding function $f_\Phi$ is the miniUNet model excluding the last $1\times1$ convolutional layer, with this, the prototypes are $32$-dimensional.

\subsection{Evaluation Protocol}\label{sec:protocol}

\subsubsection{Baselines}\label{sec:baselines}

We use two baselines for comparison with our approaches: 1) \textit{From Scratch} and 2) \textit{Fine-Tuning}. Given our few-shot semantic segmentation problem parameters in the form of the set of segmentation task $\{\St_1 , \St_2 , \dots, \St_n \}$, and a few-shot task $\F$, we define our baselines as follow:
\par\noindent\textbf{From Scratch}: Given our miniUNet network, we perform a simple supervised train with the Few-shot task support set ($\D^{sup}_{\mathcal{F}}$). We use the same Cross-Entropy loss ignoring unlabeled pixels as our cost function (Equation~\ref{eq:selective_CE}). The Adam optimizer~\citep{kingma2017adam} was used, with the same parameter used in the training of our methods.
\par\noindent\textbf{Fine-tuning}: We use the miniUNet architecture. We choose one task $\mathcal{S}_i$ from our tasks set, and perform a supervised train with the $\D_{\mathcal{S}_i}^{sup}$. Once finished the training on $\mathcal{S}_i$, we perform the fine-tune (a supervised training) using the $\D^{sup}_{\mathcal{F}}$ set. Again, the same Cross-Entropy loss function (Equation~\ref{eq:selective_CE}) is used with the same parametrized Adam optimizer.

    
    
We choose to not present the Guided Nets \cite{rakelly2018conditional} as a baseline in this work. The use of pre-trained CNN as features extractors seen to be essential to the efficiency of the model, and did not translate well to our evaluate scenarios (medical and remote sensing). To the best of our efforts the model was not able to converge to a usable model with our Meta-Datasets. Thus, it did not seem fair to compare the Guided Nets to our approach.

\subsubsection{Protocol and Metrics}\label{sec:ev_protocol}

In order to assess the performance of our methods in a certain setting, we employ a Leave-One-Task-Out methodology. That is, all but the pair $(dataset, class)$ chosen as the Few-shot task ($\mathcal{F}$) are used in the Meta-Dataset, reserving $\mathcal{F}$ for the tuning/testing phase. This strategy serves to simultaneously hide the target task from the meta-training while also allowing the experiments to evaluate the proposed algorithm and baselines in a myriad of scenarios. Moreover, we divide our tasks into two groups to perform the experiments: (I) \textbf{Medical Tasks}: all the medical datasets and their classes are used for these tasks, totaling $13$ tasks;
(II) \textbf{Remote Sensing Tasks}: we used rural datasets for these tasks. There are $5$ tasks in total ($4$ from the Brazilian Coffee and $1$ from the Orange Orchards dataset).

For each method, we used a different number of epochs in each of their training phases. In table~\ref{tab:epoch_configs}, we show these numbers that differ mostly due to training time. The Remote Sensing datasets are, in general, larger than the Medical datasets, and this made the training process (that includes validation) more time-consuming.
We use the Adam optimizer~\citep{kingma2017adam} with learning rate $0.001$, weight decay $0.0005$, and momentum $0.9$. Our batch size was set to $5$. The number of tasks sampled for the inner loop of \mamlmethod and \protomethod was set to $6$ in Medical experiments and $4$ in Remote Sensing experiments due to memory constraints, in general, and the total number of tasks in Remote Sensing experiments. 


\begin{table}[h]
    \centering
    \caption{Number of epochs for training the methods in different experiments.}
    \label{tab:epoch_configs}
    \begin{adjustbox}{width=\columnwidth}
        \begin{tabular}{@{}lcccc@{}}
            \toprule
            \multicolumn{1}{c}{\textbf{Method}} & \multicolumn{2}{c}{\textbf{Medical Experiments}} & \multicolumn{2}{c}{\textbf{Remote Sensing Experiments}} \\ \midrule
             \textit{Total Epochs} & \textbf{Pre/Meta-Training} & \textbf{Tuning} & \textbf{Pre/Meta-Training} & \textbf{Tuning} \\
            \textbf{\mamlmethod} & 2000 & 80 & 200 & 40 \\
            \textbf{\protomethod} & 2000 & - & 200 & - \\
            \textbf{Fine-Tuning} & 200 & 80 & 100 & 80 \\
            \textbf{From Scratch} & - & 80 & - & 100 \\ \bottomrule
        \end{tabular}
    \end{adjustbox}
\end{table}

We use a 5-fold cross-validation protocol in the experiments. Each dataset had a training and validation partition for each fold. Once fix the experiment fold, the support sets for the tasks are obtained from the training partition of the dataset, while the query sets are the entire validation partition. 

All images and labels are resized to $256\times256$ for remote sensing images and $128\times128$ for medical images prior to being fed to the models. This was due to our infrastructure limitations and done to standardize the input size and minimize the computational cost of the methods, especially the memory footprint of \mamlmethod method, due to the computation of second derivatives, on high-dimensional outputs. 

The metric within a fold is computed for all images in the query set according to the dense labels, and is averaged in relation to the images in that fold. 
The metric used is the Jaccard score (or Intersection over Union -- IoU) of the validation images, a common metric for semantic segmentation.


%% file: experiments/datasets.tex

\begin{figure*}[t!]
    \centering
    \includegraphics[width=\textwidth]{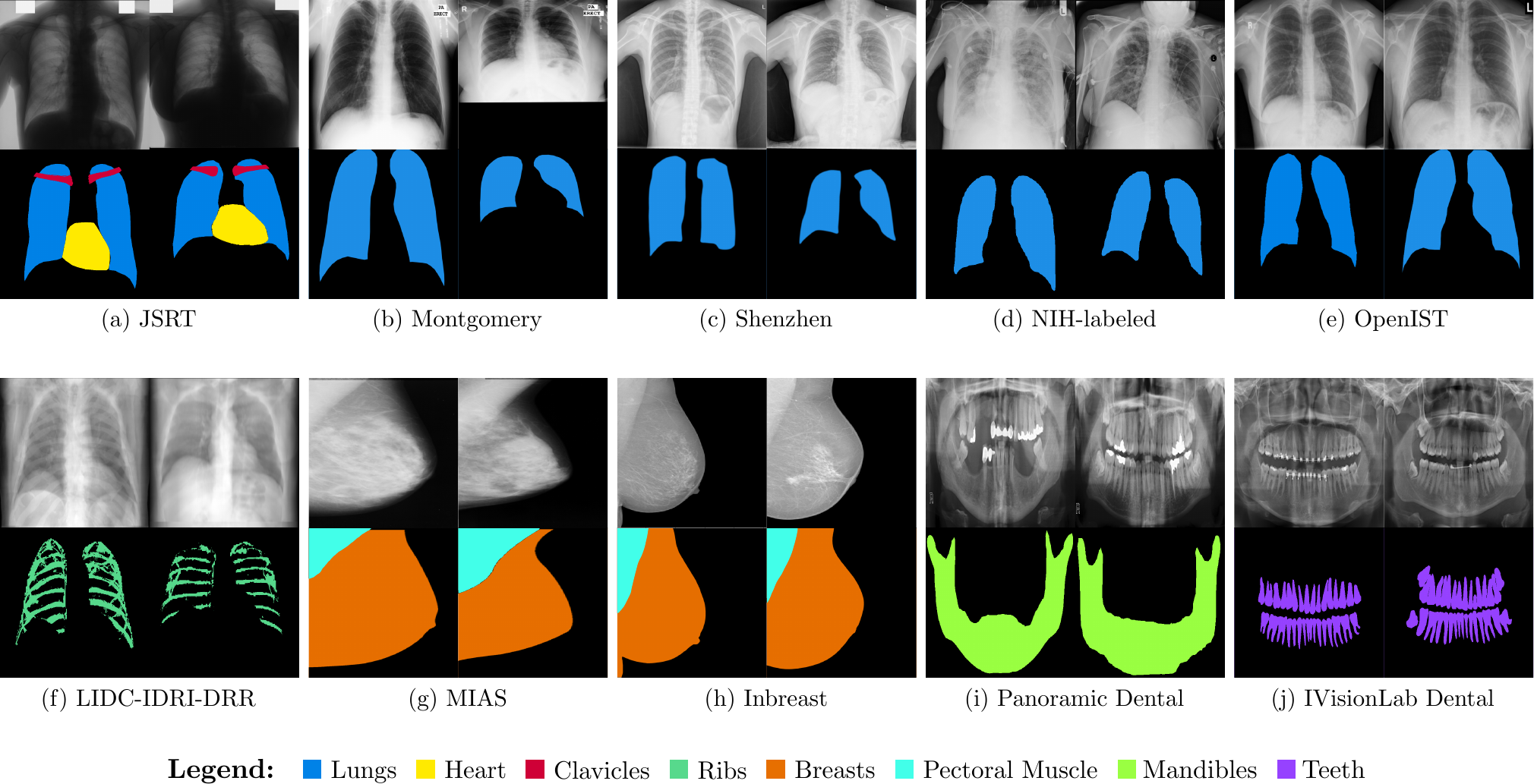}
    \caption{Examples of biomedical imaging datasets included in our medical meta-dataset.}
    \label{fig:med_examples}
\end{figure*}

\subsection{Datasets}\label{sec:datasets}

We design experiments to evaluate the proposed methods in medical and remote sensing applications for semantic segmentation. These areas have some similarities that contrast them from others. Their images are rather distinct from common RGB images taken with surveillance or cellphone cameras, for instance. 
This hinders the knowledge transfer from other generic domains or the use of pre-trained models on large datasets such as ImageNet. Another common aspect of these two areas is the limitation of availability of images due to multiple factors. Medical image datasets have to face privacy and ethical concerns, also requiring a highly specialized radiologist to provide precise annotations. In remote sensing the annotation is extremely laborious and sometimes unfeasible since it typically requires that a specialist collect information from large geographical areas, maybe even requiring visits to the site for producing ground truths for these data.

\subsubsection{Medical Imaging Datasets}\label{sec:med_datasets}

We use a total of ten medical datasets in our experiments (Figure~\ref{fig:med_examples}). Of these datasets, six are Chest X-Ray datasets (CRX): JSRT \cite{JSRTshiraishi2000development} with labels for lungs, heart and clavicles; the Montgomery/Shenzhen sets \cite{jaeger2014two}, an annotated subset of Chest X-Ray 8 \cite{NIHwang2017chestx} by \citet{NIHtang2019xlsor} referred to as NIH-labeled,  OpenIST\footnote{\url{https://github.com/pi-null-mezon/OpenIST}} with labels for lung segmentation, and the LIDC-IDRI-DRR dataset~\citep{LIDColiveira20203d}, with generated ribs annotations. We include two Mammographic X-Ray (MRX) image sets, namely INbreast \cite{moreira2012inbreast} and MIAS \cite{MIASsuckling1994mammographic}, with labels for breast region and pectoral muscle segmentation. Also, two Dental X-Ray (DRX) datasets are included: Panoramic X-Ray~\cite{PANORAMICabdi2015automatic} with labels for the inferior mandible and IVisionLab~\cite{IVISIONsilva2018automatic} annotated for teeth segmentation.

\subsubsection{Remote Sensing Datasets}\label{sec:rs_datasets}
\begin{figure}[h]
    \centering
    \includegraphics[width=\columnwidth]{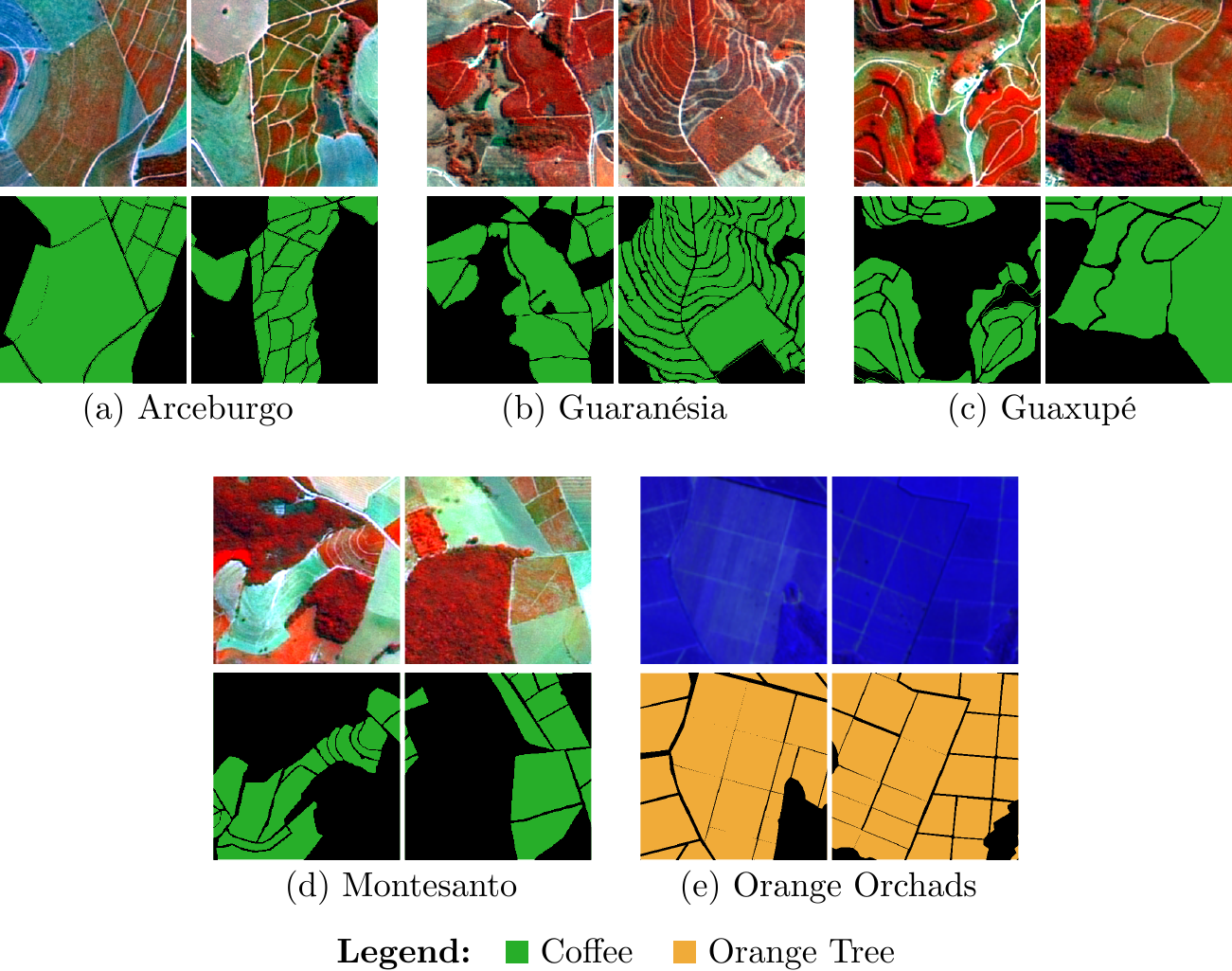}
    \caption{Examples from the Brazilian Coffee and Orange Orchards Datasets.}
    \label{fig:coffee_example}
\end{figure}

The Remote Sensing meta-dataset is composed of rural scenes for crop segmentation (Figure~\ref{fig:coffee_example}). More specifically, we use the Brazilian Coffee dataset, composed of images of 4 municipalities -- namely, Arceburgo, Guaran\'{e}sia, Guaxup\'{e} and Montesanto -- with pixel-level annotations for coffee crops regions, as well as the Orange Orchards (Ubirajara county, Brazil) dataset, with annotations for orange crop regions. Both datasets will be made public available upon the acceptance of this work.

Further description of all datasets are presented in the supplementary material.

%% file: results/results.tex
\section{Results and Discussion}
\label{chap:results}

In this section, we present and discuss the results of our experiments. Section~\ref{sec:sparse_v_dense} shows a comparison of the results of the proposed methods and baselines using multiple sparse annotations and their densely annotated counterparts. Section~\ref{sec:med_results} focuses on the medical imaging datasets, while Section~\ref{sec:agriculture_results} describes the results obtained from remote sensing data. At last, in Section~\ref{sec:sparse_comparison} we evaluate different sparse annotation styles regarding the number of user inputs and segmentation performance. 

\subsection{Few-shot Semantic Segmentation: Sparse vs Dense labels}\label{sec:sparse_v_dense}

In this section, we present the results of our methods in multiple few-shot tasks in the Medical and Remote Sensing scenarios.
We evaluated different number of shots and parameters for each type of sparse annotation.
In Sections~\ref{sec:med_results} and~\ref{sec:agriculture_results}, we present the results grouped in plots organized by sparse annotation type and number of shots. Dashed lines in the graphs represent the scores of the methods trained with dense annotations.

\subsubsection{Medical Tasks}\label{sec:med_results}

Analyzing the results in the CRX tasks, two trends can be easily seen. First, an obvious insight that holds for most methods and scenarios is that better scores are obtained with more data. Larger support sets (higher $k$-shots) and more sparsely annotated pixels result in better performance for the algorithms. A second observed result is that \mamlmethod surpassed the performances of \protomethod and baselines in tasks with a larger domain shift to other tasks in the meta-dataset. This was observed mainly in the \textit{JSRT Lungs} (Figure~\ref{fig:jsrt_lungs_results}) and the \textit{JSRT Heart} (Figure~\ref{fig:jsrt_heart_results}) experiments, as the JSRT dataset is visually the most distinct of the CRX datasets. Additionally, the \textit{Heart} class is annotated only in this dataset, resulting in a large domain shift in the semantic space for this task in comparison to the other tasks used in the meta-training.

\newcommand{\currprop}{0.85\textwidth}

\begin{figure*}[h]
    \centering
    \includegraphics[width=\currprop]{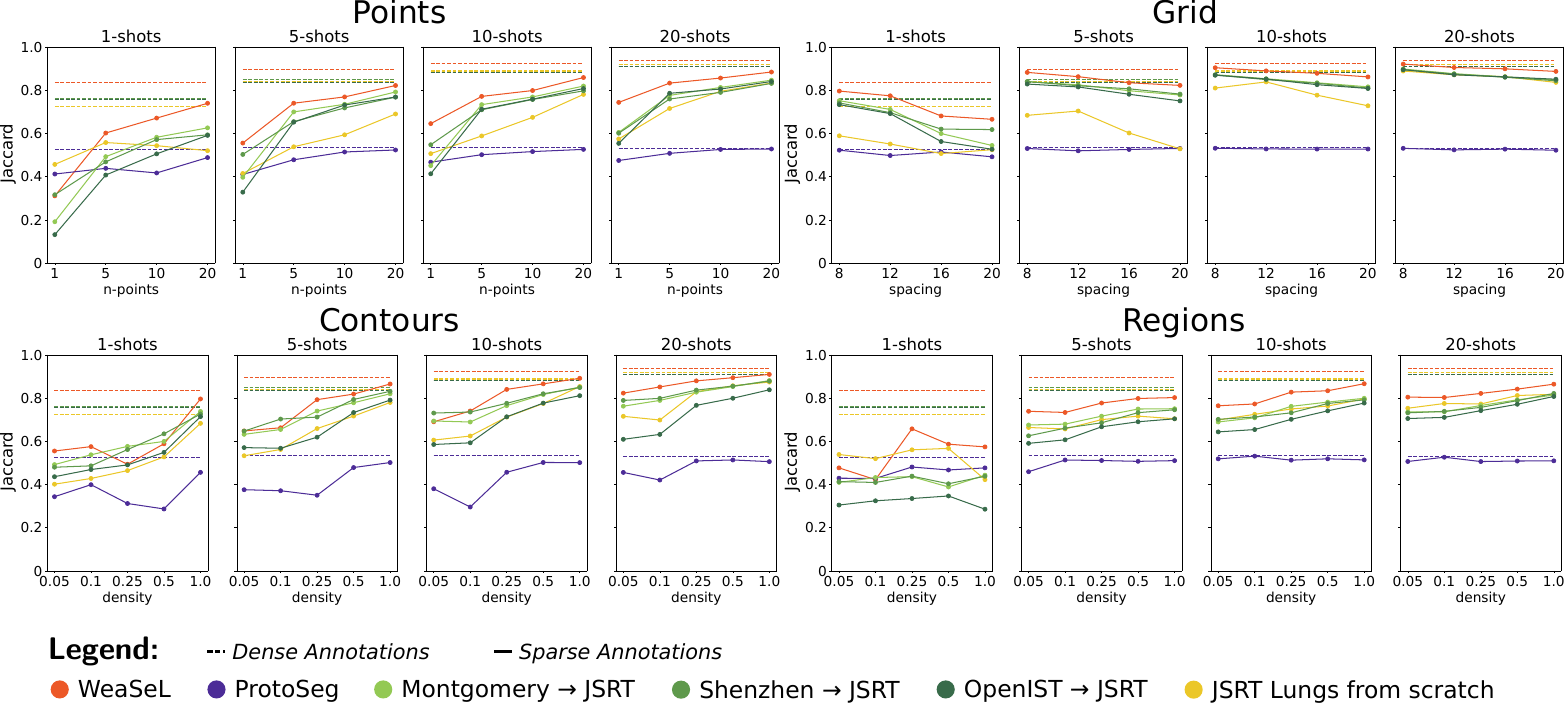}
    \caption{Jaccard score of experiments with \textit{JSRT Lungs} task.}
    \label{fig:jsrt_lungs_results}
\end{figure*}

\begin{figure*}[h]
    \centering
    \includegraphics[width=\currprop]{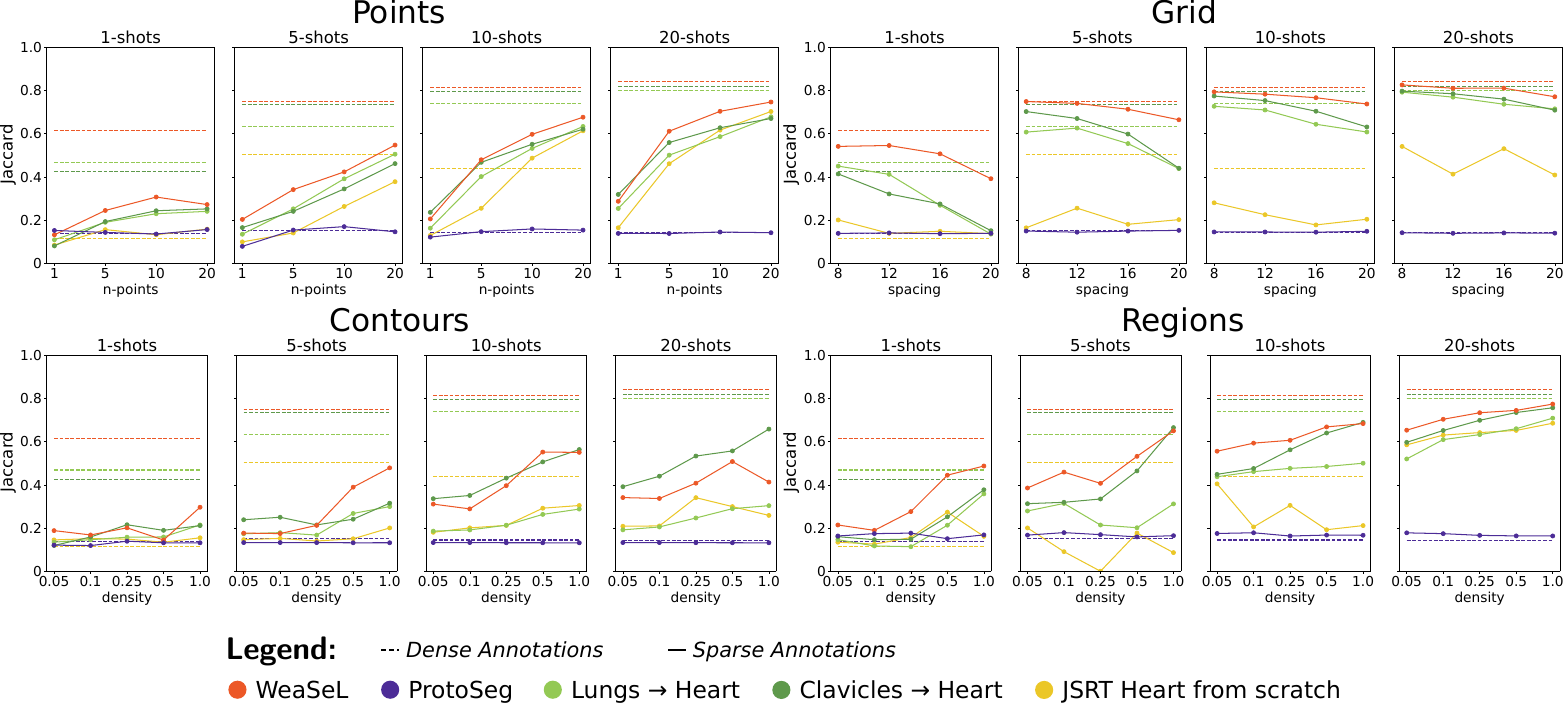}
    \caption{Jaccard score of experiments with \textit{JSRT Heart} task.}
    \label{fig:jsrt_heart_results}
\end{figure*}

\begin{figure*}[h]
    \centering
    \includegraphics[width=\currprop]{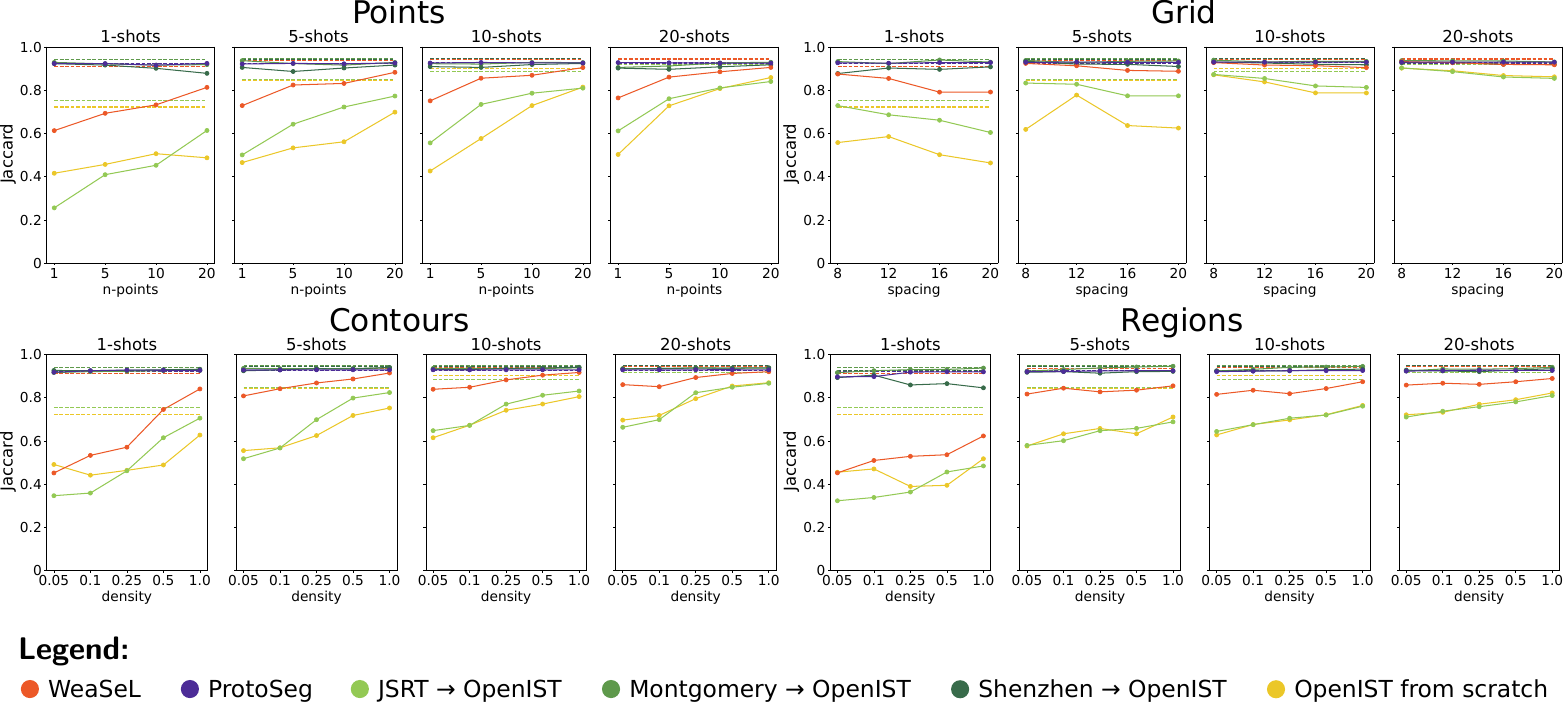}
    \caption{Jaccard score of experiments with \textit{OpenIST Lungs} task.}
    \label{fig:openist_lungs_results}
\end{figure*}

For the remaining tasks, we observe that either the \protomethod method or some fine-tuning baseline is the best performer. Since some datasets are visually similar, fine-tuning for a task from a model trained in a similar dataset is a known viable solution that works well in these cases. 
Fine-tuning from similar tasks (e.g., OpenIST, Montgomery, or Shenzhen for lung segmentation) yields the best Jaccard scores in most cases, as exemplified in Figure~\ref{fig:openist_lungs_results} for the OpenIST dataset. Also, we observe that the \protomethod is consistently comparable to these fine-tuned baselines. The same plots (as in Figure~\ref{fig:openist_lungs_results}) for Montgomery and Shenzhen tasks, omitted in this text due to size constraints, can be found in the supplementary material.

\begin{figure*}[h]
    \centering
    \includegraphics[width=\currprop]{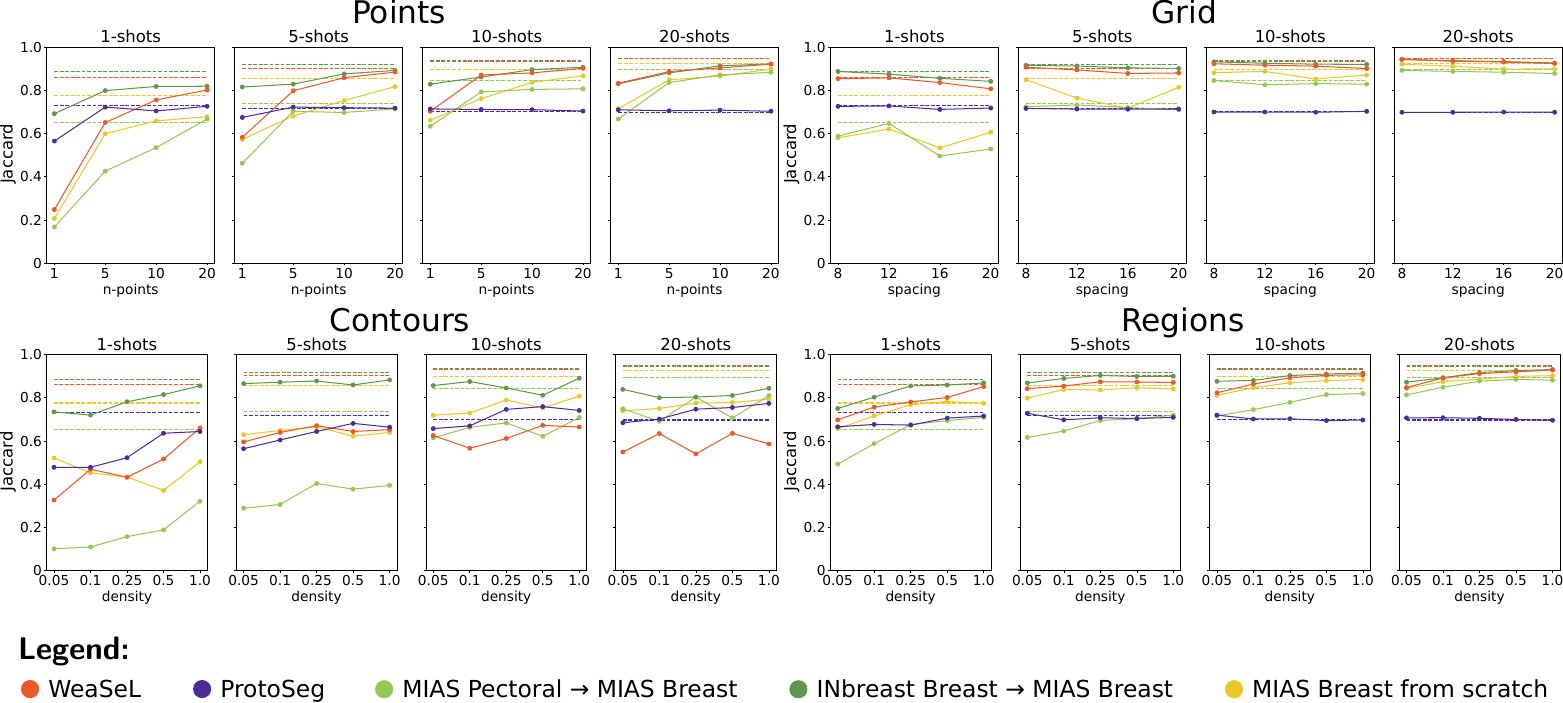}
    \caption{Jaccard score of experiments with \textit{MIAS Breast} task.}
    \label{fig:mias_breast_results}
\end{figure*}

MRX and DXR tasks present similar tendencies as the ones observed in the CRX datasets. Again, fine-tuning from similar tasks appears as a solid solution, with \mamlmethod obtaining comparable results to the baselines in most cases. In the \textit{MIAS Breast} task (Figure~\ref{fig:mias_breast_results}), fine-tuning from the \textit{INbreast Breast} proved to be the best method, mainly due to having the same semantic space on the source and target domains, closely followed by \mamlmethod in most scenarios.

\begin{figure*}[h]
    \centering
    \includegraphics[width=\currprop]{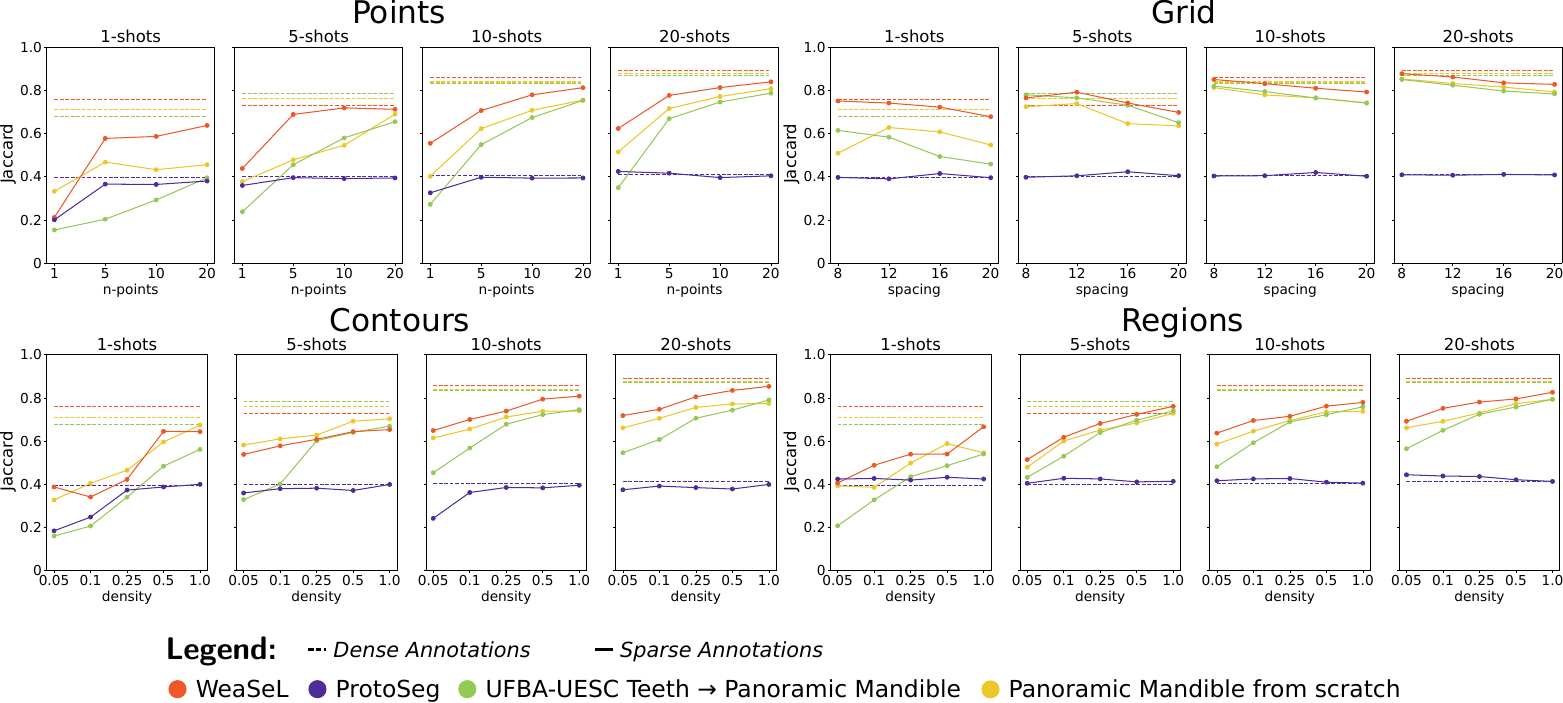}
    \caption{Jaccard score of experiments with \textit{Panoramic Mandible} task.}
    \label{fig:panoramic_mandible_results}
\end{figure*}

Two DXR datasets are included in the meta-dataset in our experiments, assuring that in experiments with one DXR dataset as target, the other one is always used for the pretraining. However, the \textit{Panoramic} dataset is labeled for mandibles while \textit{IVisionLab} data are labeled for teeth, hence never sharing the same label space. In this scenario, without a task with similar semantic space to fine-tune from, the \mamlmethod yields the best performance in the segmentation tasks, achieving the highest scores in the majority of experiments with both dense and sparse annotations. This can be observed in Figure~\ref{fig:panoramic_mandible_results} for the \textit{Panoramic Mandible} task. Being the most distinct tasks, even the from scratch baseline yields comparable results to the more complex alternatives, in some cases even achieving the best results. \protomethod underperforms by a large margin in comparison to the other methods in \textit{Panoramic Mandible}, which can be explained by the low prevalence of DXR data in the meta-dataset used for meta-training and by the large semantic space domain shift even among DXR datasets. 

\subsubsection{Remote Sensing Tasks}\label{sec:agriculture_results}

\begin{figure*}[h]
    \centering
    \includegraphics[width=\currprop]{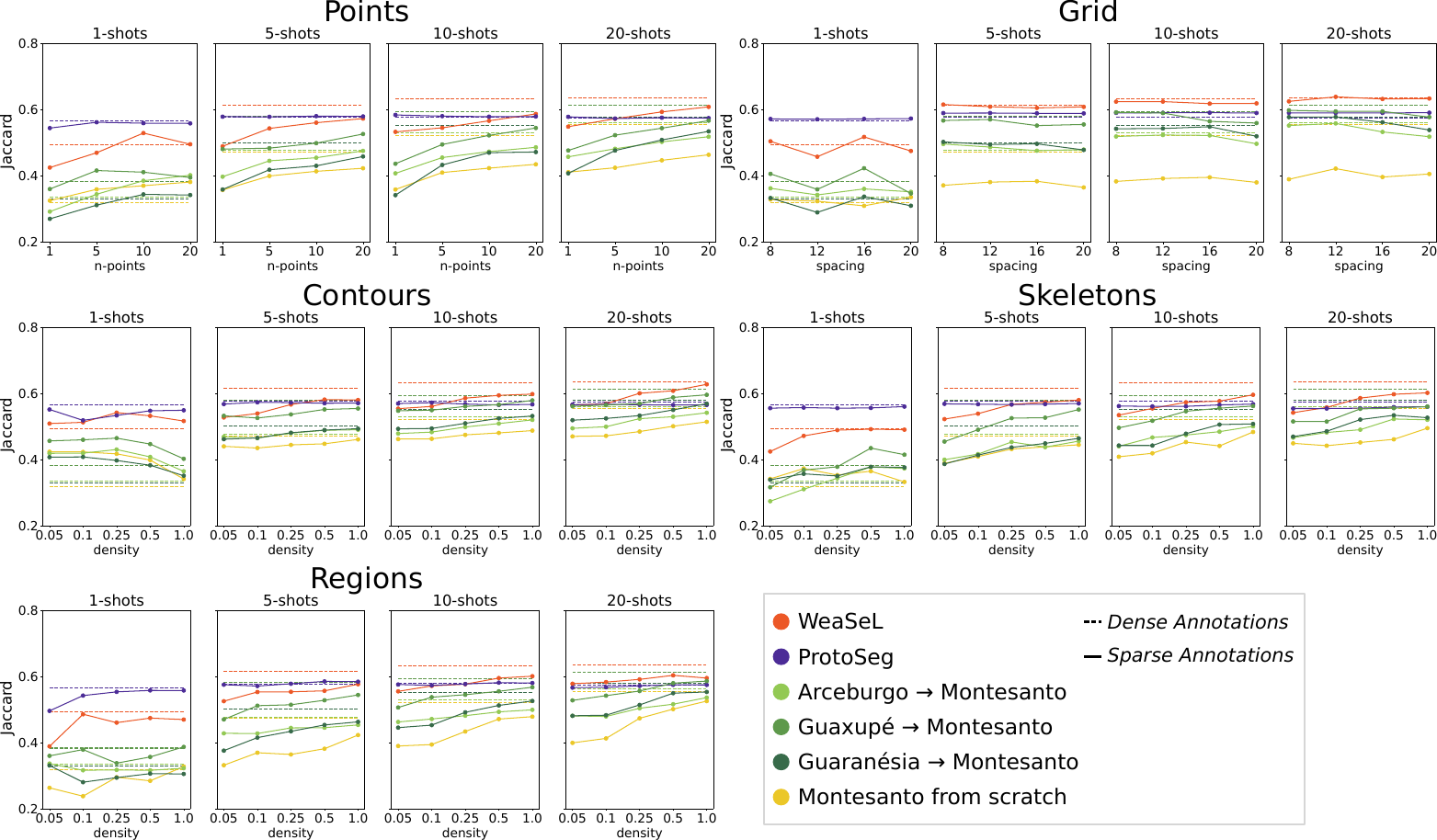}
    \caption{Jaccard score of experiments with \textit{Montesanto Coffee} task.}
    \label{fig:mon_results}
\end{figure*}

\begin{figure*}[h]
    \centering
    \includegraphics[width=\currprop]{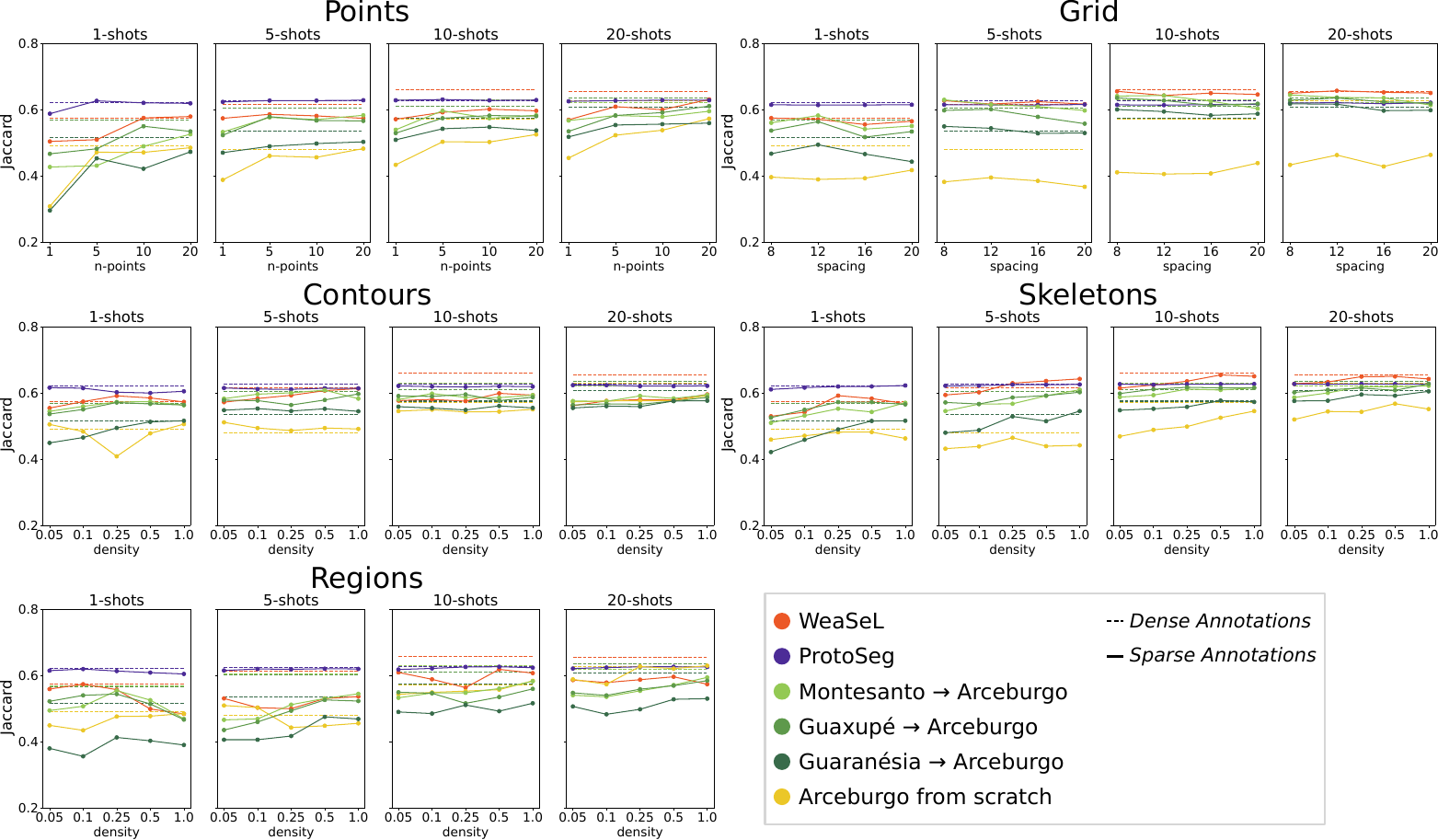}
    \caption{Jaccard score of experiments with \textit{Arceburgo Coffee} task.}
    \label{fig:arc_results}
\end{figure*}

\begin{figure*}[h!]
    \centering
    \includegraphics[width=\currprop]{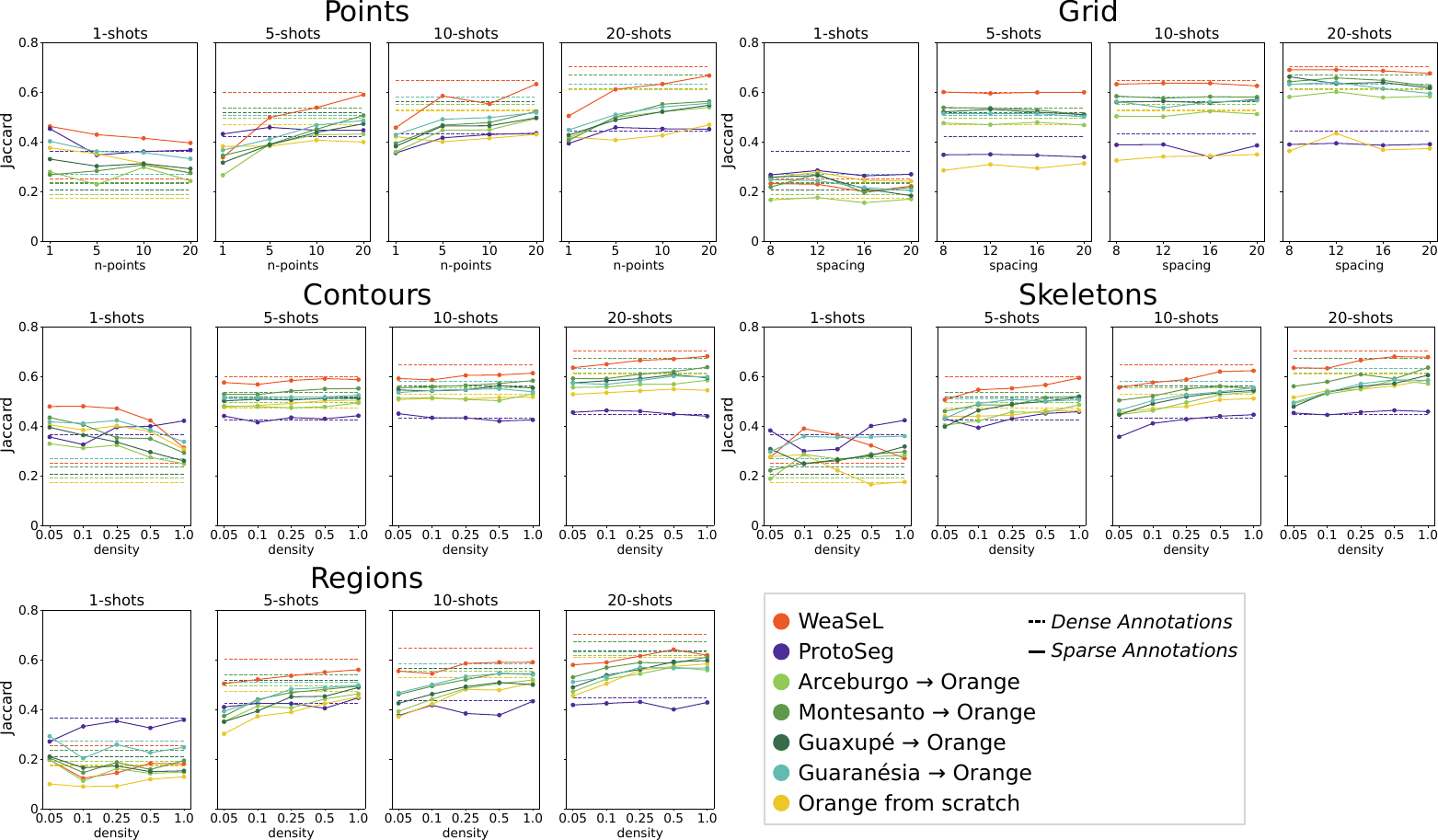}
    \caption{Jaccard score of experiments with \textit{Orange Orchard} task.}
    \label{fig:lrj_results}
\end{figure*}

In general, all remote sensing tasks proved to be considerably harder than the medical ones. Overall, no method achieved a Jaccard score above $0.8$ in any of the evaluated tasks, not even when using dense labels. Figures~\ref{fig:mon_results} and~\ref{fig:arc_results} depict the results for the \textit{Montesanto Coffee} and \textit{Arceburgo Coffee} tasks, from the Brazilian Coffee dataset, while Figure~\ref{fig:lrj_results} shows results for the \textit{Orange Orchard} task. One can easily observe that the \mamlmethod method consistently outperforms fine-tuning and from scratch baselines, especially in configurations with few data (1-shot tasks). Although having the same label space, the coffee segmentation presents an intrinsic domain shift across the 4 different counties in the dataset. This is due to distinct geographical features, coffee crop cycles and/or plantation methods, explaining why simple fine-tuning is not always the best solution for coffee crop segmentation.

\protomethod had consistent results in most agricultural tasks. For the \textit{Coffee} tasks, it generally obtained Jaccard scores around $0.6$, while the performance in the \textit{Orange} task revolved around $0.4$. In a tendency similar to the Medical experiments, \protomethod seems to benefit from very related/similar tasks, particularly regarding the semantic space. 
When choosing \textit{Orange Orchard} as the target task, only the \textit{Coffee} tasks were available to be used in training, explaining its lower performance in comparison to the \textit{Coffee} datasets. 

\subsection{Sparse Label Efficiency Comparison}
\label{sec:sparse_comparison}


In this section, we present results for three types of sparse annotations: \textit{Points}, \textit{Grid}, and \textit{Regions}. \textit{Contour} and \textit{Skeleton} annotations are not evaluated due to our methods to generate them. We define the number of user inputs for a type of annotation as the number of interactions an annotator would have to perform to annotate the image sparsely using said type. 

For a single image, the number of inputs for a $n$-point \textit{Points} annotation  is $2n$: the $n$ positive and $n$ negative pixels selected. For the \textit{Grid} annotation, the number of inputs is the total positive labeled pixels in the grid since they are initially assumed to be negative, and the user picks the positive ones. As for the \textit{Regions} annotation, the number of inputs is defined as the total regions selected, independent of being positive or negative. After the number of inputs for a single image is computed, the values are summed for all images in the support set of the $k$-shot task. Then, the number of inputs of the $k$-shot are averaged across the five folds.

Figures~\ref{fig:inp_jsrt_results} and \ref{fig:inp_montesanto_results} present results label efficiency plots for \textit{JSRT Lungs} and \textit{Montesanto Coffee}, respectively. 
We observe that, as seen in Section~\ref{sec:sparse_v_dense}, the \mamlmethod method overall performs better with more data, that increases with the number of user inputs. We also clearly see how the \protomethod method is almost indifferent to the sparsity and quantity of annotations by having a low deviation score in the presented tasks. 
For the \textit{JSRT Lungs} task (Figure~\ref{fig:inp_jsrt_results}), and Medical tasks, in general, we see that the \textit{Grid} annotation usually achieve a higher score for the same number of user inputs as the other types of annotation. On the other hand, the \textit{Region} annotation is commonly the best annotation type for the Remote Sensing tasks, having higher scores with the same number of inputs. The \textit{Points} annotation is, at most times, the worse performer. This was expected since with the same number of inputs, this type of annotation will have fewer labeled pixels in total than the other types.

\renewcommand{\currprop}{0.95\columnwidth}

\begin{figure}[h]
    \centering
    \includegraphics[width=\currprop]{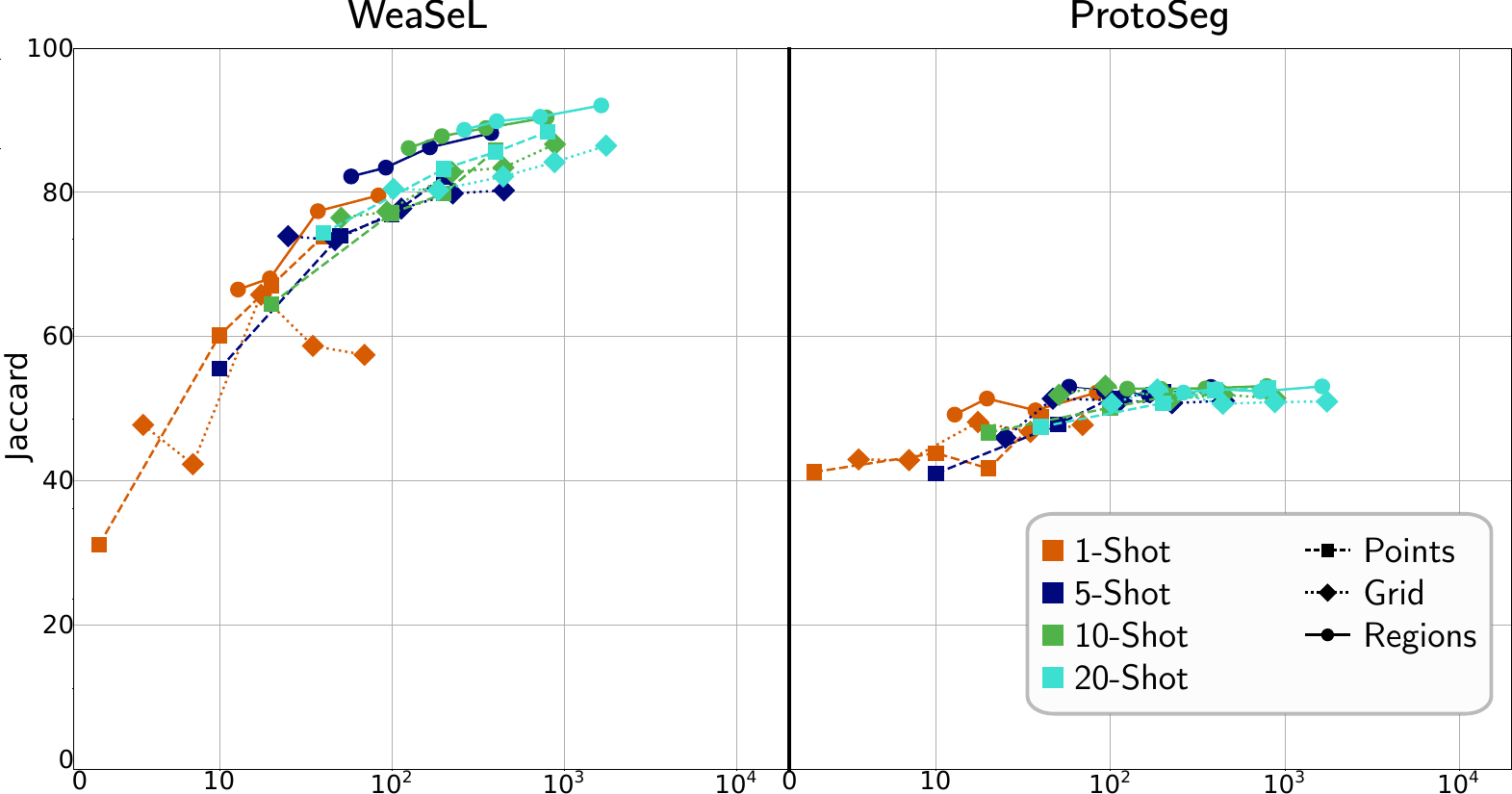}
    \caption{Number of user inputs versus Jaccard score in the \textit{JSRT Lungs} task.}
    \label{fig:inp_jsrt_results}
\end{figure}

\begin{figure}[h]
    \centering
    \includegraphics[width=\currprop]{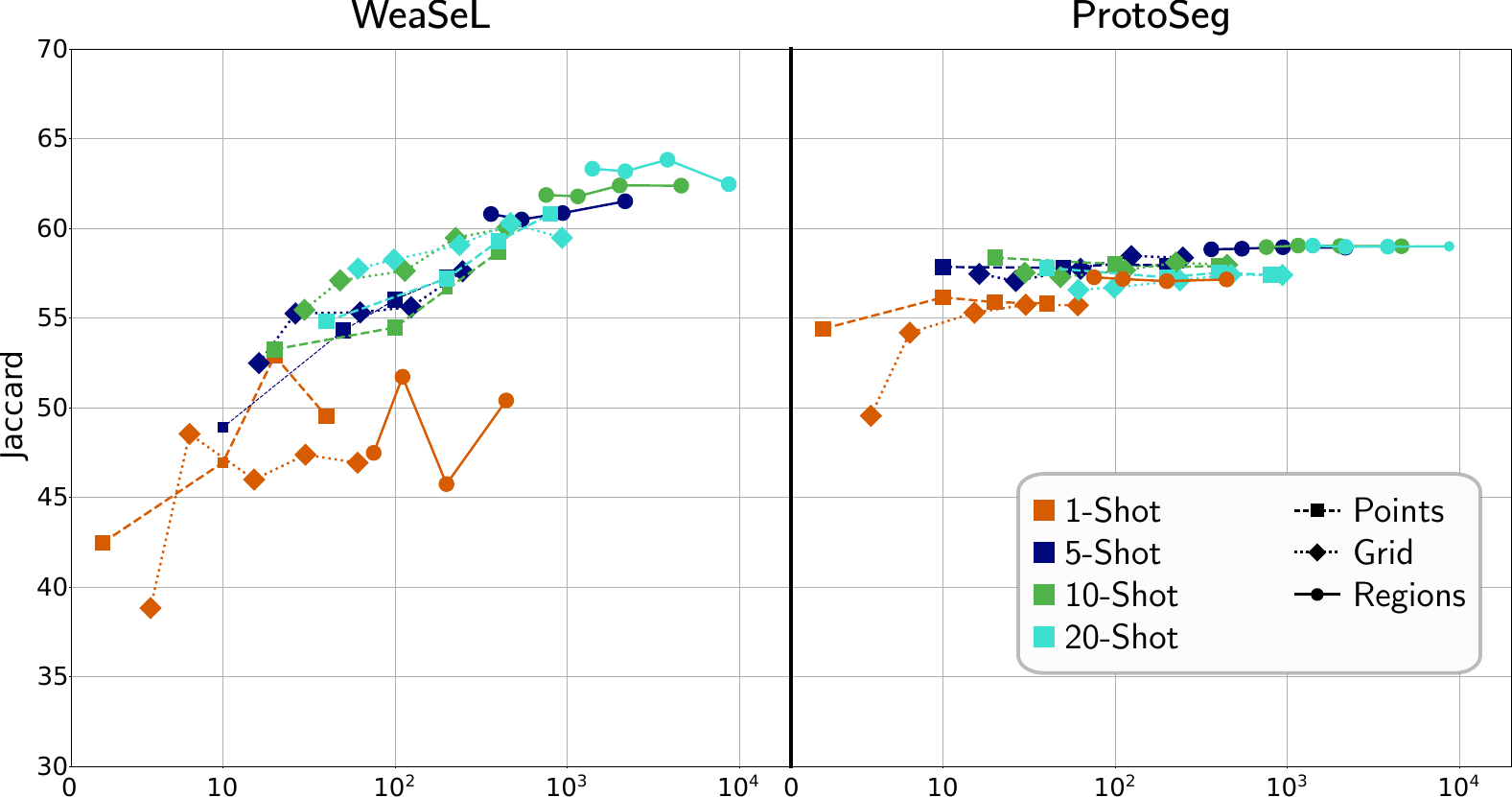}
    \caption{Number of user inputs versus Jaccard score in the \textit{Montesanto Coffee} task.}
    \label{fig:inp_montesanto_results}
\end{figure}



By comparing these results and the ones of the previous section (Section~\ref{sec:sparse_v_dense}) we can draw some discussions. The \textit{Grid} annotation is a solid annotation type that can lead to good results and usually is one of the best types for Medical cases. However, this is the most user-consuming type, requiring a larger number of user inputs. The \textit{Regions} annotation is also a solid option for annotations. When the superpixel segmentation produces clean regions easier to be labeled, this type can make annotation simpler and quicker and produce precise models, especially for Remote Sensing tasks. The \textit{Points} annotation 
requires less from the user. It does not produce the most optimal models but can lead to comparable results requiring far fewer inputs. Also, this annotation guarantees a balanced number of pixels samples for each class in training, making optimization of the models easier. The other two types of annotations, \textit{Contours} and \textit{Skeletons}, appear as valid options as well. The way we designed the process of generating these annotations made it difficult to translate to a countable user input, which is why these types are not compared in this section. However, the results presented in Section~\ref{sec:sparse_v_dense} show that \textit{Contours} and \textit{Skeletons} annotations are suitable styles, specially \textit{Contours} in the Medical tasks and \textit{Skeletons} in Remote Sensing tasks.

%% file: conclusion/conclusion.tex
\section{Conclusion and Future Works}\label{chap:conclusion}

In this work, we proposed one method (\cprotomethod) to the problem of weakly supervised few-shot semantic segmentation, also conducting extensive experiments on similar previously proposed method (\mamlmethod \cite{gama2021weakly}). Despite being common in shallow interactive segmentation methods, few-shot segmentation from sparse labels is still not fully integrated with the advances in computer vision brought by Deep Learning. 
We evaluated our two meta-learning methods in a large number of experiments to verify their generalization capabilities in multiple image modalities, number of shots, annotation types and label densities. We chose to focus the experiments in two areas that can benefit from few-shot sparse labeled semantic segmentation: medical imaging and remote sensing. 

\mamlmethod \cite{gama2021weakly}, obtained promising results, mainly in scenarios with a large domain shift between the target and source tasks.
The proposed \protomethod method yielded reliable segmentation predictions in the cases wherein there are multiple closely related source datasets, as good results from \protomethod appear to be correlated to the availability of similar tasks during training.
The five annotation types evaluated in our experiments --- \textit{Points}, \textit{Grid}, \textit{Contours}, \textit{Skeletons}, and \textit{Regions} --- have their own pros and cons. The \textit{Grid} annotation proved to be highly reliable and produce some of the best results, even though it often requires more user intervention. \textit{Region} annotations can be a more efficient option, but its usefulness is highly affected by the performance of the superpixel segmentation algorithm. \textit{Points} annotations are the less user demanding, but it also yields the greater gaps for dense annotation scores. \textit{Contours} and \textit{Skeletons} appear as valid options for medical imaging and remote sensing tasks, respectively. However more experiments must be conducted to confirm their efficiency in comparison to the other label modalities.


For future works, we intend to investigate adding spacial reasoning into the segmentation predictions in order to account for the location and feature representations of a given pixel in comparison to annotated pixels. Additionally, further experiments in other medical imaging (e.g. other 2D x-ray exams, volumetric images, etc) and remote sensing (e.g. urban segmentation tasks) will be conducted using both \protomethod and \cmamlmethod. At last, our team shall investigate the need for real annotations at all during the meta-training phase. Instead, we plan to replace the sparse masks for organs and crops by automatically generated weakly supervised masks of regions obtained by shallow unsupervised segmentation algorithms.
